\documentclass[11pt,3p,review,authoryear]{elsarticle}
\usepackage{siunitx}
\usepackage{natbib}
\usepackage{amsmath}
\usepackage{url}
\usepackage{xcolor}
\usepackage{amsfonts}
\usepackage{amsmath}
\usepackage[titletoc,page]{appendix}
\usepackage{float}
\usepackage{todonotes}
\usepackage{blindtext}
\usepackage{subfig}
\usepackage[nameinlink,capitalise]{cleveref}

\usepackage[export]{adjustbox}

\NewDocumentCommand{\anote}{}{\makebox[0pt][l]{$^*$}}

\begin{document}

\begin{frontmatter}

\title{LoMEF: A Framework to Produce Local Explanations for Global Model Time Series Forecasts}



\author[monash1]{Dilini~Rajapaksha \corref{cor1}}
\author[monash1]{Christoph~Bergmeir}
\author[monash2]{Rob J~Hyndman}

\address{Dilini.Rajapakshahewaranasinghage@monash.edu, Christoph.Bergmeir@monash.edu, Rob.Hyndman@monash.edu}
\address[monash1]{Department of Data Science \& Artificial Intelligence, Monash University, Clayton VIC 3800, Australia.}
\address[monash2]{Department of Econometrics \& Business Statistics, Monash University, Clayton VIC 3800, Australia.}

\cortext[cor1]{Postal Address: Faculty of Information Technology, P.O. Box 63 Monash University, Victoria 3800, Australia. E-mail address: Dilini.Rajapakshahewaranasinghage@monash.edu}


\begin{abstract}
Global Forecasting Models (GFM) that are trained across a set of multiple time series have shown superior results in many forecasting competitions and real-world applications compared with univariate forecasting approaches. One aspect of the popularity of statistical forecasting models such as ETS and ARIMA is their relative simplicity and interpretability (in terms of relevant lags, trend, seasonality, and others), while GFMs typically lack interpretability, especially towards particular time series. This reduces the trust and confidence of the stakeholders when making decisions based on the forecasts without being able to understand the predictions. To mitigate this problem, in this work, we propose a novel local model-agnostic interpretability approach to explain the forecasts from GFMs. We train simpler univariate surrogate models that are considered interpretable (e.g., ETS) on the predictions of the GFM on samples within a neighbourhood that we obtain through bootstrapping or straightforwardly as the one-step-ahead global black-box model forecasts of the time series which needs to be explained. After, we evaluate the explanations for the forecasts of the global models in both qualitative and quantitative aspects such as accuracy, fidelity, stability and comprehensibility, and are able to show the benefits of our approach.
\end{abstract}


\end{frontmatter}

\section{Introduction}
We are living in an era of big data. Big data in the field of time series forecasting does not necessarily mean an existence of lots of data points in a single time series. There are many applications and databases which consist of large quantities of time series from similar sources in the same domain. In this context, forecasting based on traditional univariate time series forecasting procedures leaves great potential untapped to provide accurate forecasts.
To address this issue, many works have recently explored the capabilities of global forecasting models~\citep[GFM,][]{januschowski2020criteria} that are trained across all available time series. Most notably the M4 forecasting competition~\citep{makridakis2018statistical} was won by a particular type of globally trained Recurrent Neural Network (RNN)~\citep{smyl2020hybrid}, the M5 competition~\citep{makridakis2020m5} was dominated by globally trained LightGBM models, earlier Kaggle competitions in this space were also dominated by globally trained models~\citep{bojer2020kaggle}, and companies like Amazon~\citep{salinas2020deepar}, Walmart~\citep{bandara2019sales}, Uber~\citep{laptev2017time}, and others are relying on such models.

However, the premise existing for decades in the forecasting space (most notably as a conclusion of competitions such as the M and M3 competitions \citep{makridakis2000m3}) that simple models are oftentimes as accurate as more complex ones, has been rendered obsolete by such GFMs, allowing them to be considerably more complex while maintaining good generalisation and extrapolation capabilities~\citep{montero2020principles}. In this sense, forecasting is following other areas such as medicine where complex machine learning models often outperform in terms of accuracy the relatively simple and interpretable predictive models established in these domains \citep{nanayakkara2018characterising}. However, the machine learning models have the drawback of being black-box models that are not interpretable, leading to slow adoption as practitioners don't trust the models and are therewith hesitant to rely on them in production settings. To address this problem, in machine learning, the concept of \emph{local interpretability} has been developed to explain predictions not globally across all predictions, but for individual predictions.

Local interpretability is usually achieved by fitting an interpretable surrogate model locally in the neighbourhood of a particular instance for which we want to explain the output of a more complex black-box model, that we will call in this context the (black-box) background model. In this way, local explanations for a specific instance can be more accurate and relevant as opposed to global explanations that do not change across the dataset.
In general, to capture the generalisation behaviour of the background model, or if no training data is available, the surrogate model is trained on new instances that are generated within the neighbourhood. Options are to perturb features or interpolate them between instances from the training set. Then, predictions from the background model on these new instances are obtained and the surrogate model is trained to mimic those predictions.
One of the first works in this space is the method of Local Interpretable Model-agnostic Explanations~\citep[LIME, ][]{Ribeiro2016-sp}. 
Here, the newly generated instances are randomly perturbed around the instance to be explained and a linear model is fitted as a surrogate model to the instances in a selected neighbourhood of the prediction that needs to be explained. The linear model is assumed to be interpretable in terms of its coefficients.
Later, \citet{Lundberg2017-hc} introduced SHapley Additive exPlanation (SHAP) which is a local interpretability algorithm based on game theory that provides consistent additive explanations, for a prediction generated by the black-box model.  
Both LIME and SHAP generate feature importance based explanations that identify the features which positively or negatively influence the prediction of the background model of an instance. 
Recent literature has also focused on rule-based explainers which provide arguably more interpretable and insightful explanations than the feature importance based explanations. The most relevant works in this line are Anchor-LIME~\citep{Ribeiro2018-mc}, LOcal Rule-based Explanations~\citep[LORE, ][]{DBLP:journals/corr/abs-1805-10820}, and Local Rule-based Model Interpretability with k-optimal Associations~\citep[LoRMIkA, ][]{rajapaksha2020lormika}.

In this paper, our aim is to apply the concept of local interpretability to the forecasting domain, to provide explanations for the forecasts generated by black-box GFMs. 
As surrogate models, we use traditional local statistical models, such as Exponential Smoothing, ARIMA, and others. We assume that these models are interpretable in terms of decompositions like trend and seasonality, relevant lags, and others. Therewith, we facilitate the adoption of GFMs into practice, bridging the gap in terms of interpretability between them and the traditional, well-etablished techniques.

In the global forecasting context, we treat (the last observations of) a single time series from the set of all series as an instance to be explained. Following the local interpretability approach, we then need to define a neighbourhood and a way to sample from the neighbourhood, to obtain data on which to train a surrogate model.
Approaches to define the neighbourhood that we explore include defining a similarity measure between series, or to simply define the neighbourhood as the time series in consideration. Then, sampling from the neighbourhood can be done with a bootstrapping procedure, or in certain situations sampling can even be omitted and the series used directly. 
We evaluate our local explainer methodology empirically on five different datasets with nine different local explainers to verify the fidelity, accuracy, stability, and comprehensibility of the local explainers. We therewith show that the local explainers are viable techniques to explain global forecasting models locally.

%

The remainder of this paper is structured as follows. 
Section~\ref{sec:approach} presents our approach.
Section~\ref{sec:Experiments} discusses the experimental setup and Section~\ref{sec:results} discusses the results in both qualitative and quantitative aspects. Section~\ref{sec:conclusion} concludes the paper and discusses future work.




\section{Our Approach}
\label{sec:approach}

In this section, we first formally define the concept of local interpretability in general, and then introduce our proposed approach of local interpretability in forecasting with GFMs. Finally, we discuss in detail the GFMs and local explainers employed in this study.  

\subsection{Local interpretability in regression and classification}
\label{sec:locint}

Let $X \in \mathbb{R}^{n\times m}$ be a training dataset consisting of $n$ instances and $m$ features, and let $\boldsymbol{z}$ be an $n$ dimensional vector of corresponding target values, which can be either numerical in a regression setting or categorical in a classification setting, and let $g$ be a given black-box background model, that allows to make predictions $\hat{z}$ from $m$ dimensional input vectors $\boldsymbol{x}$, that is, $g(\boldsymbol{x})= \hat{z}$, so that an error criterion between $z$ and $\hat{z}$ is minimised. The goal of local interpretability algorithms is to find a surrogate model $e$, for example, a linear model, such that for a given instance $\boldsymbol{x}$ for which the prediction $\hat{z}$ of the background model is to be explained, $e(\boldsymbol{x}) \approx \hat{z}$.

To build $e$, we define a neighbourhood for $\boldsymbol{x}$ using a distance function, and now we can create a set of randomly generated instances that lie in the neighbourhood. We call this new set $\tilde{X}$. We then apply $g$ to all instances in $\tilde{X}$ to obtain predictions $\tilde{z}$. With this, $e$ is trained on $\tilde{X}$ to resemble $\tilde{z}$ as the target predictions. To generate $\tilde{X}$, many algorithms first filter the original training set $X$ for instances that lie in the neighbourhood of $\boldsymbol{x}$, into a set that we call $X_{\textit{filt}}$. Then, $\tilde{X}$ is generated from $X_{\textit{filt}}$ using a sampling procedure, that perturbs single instances from $X_{\textit{filt}}$ by adding noise to the features, and/or that interpolates between instances of $X_{\textit{filt}}$.

\subsection{Local interpretability in forecasting}

\begin{figure*}
	\centering
	\includegraphics[width=\textwidth]{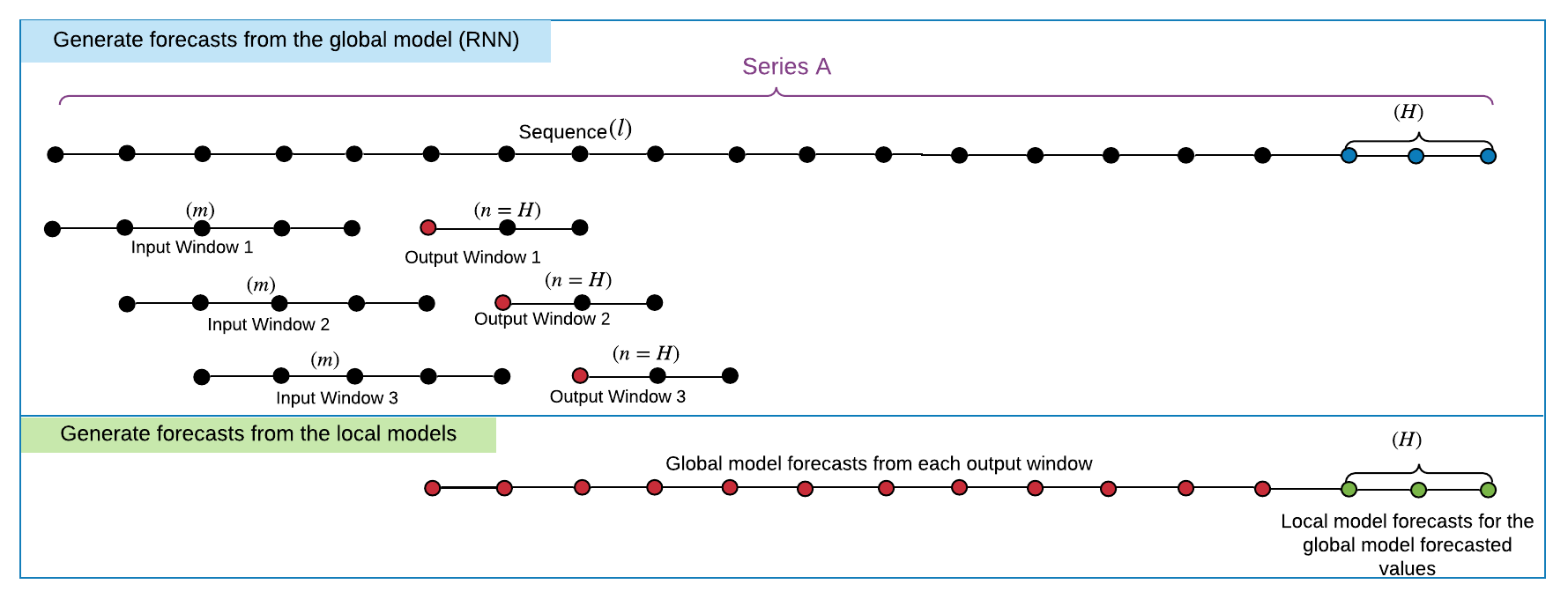}
	\caption{A diagram which shows how to extract the fit of the global model}
	\label{Fig:one_step_ahead_forecast}
\end{figure*}

\begin{figure*}[]
	\centering
	\includegraphics[width=\textwidth]{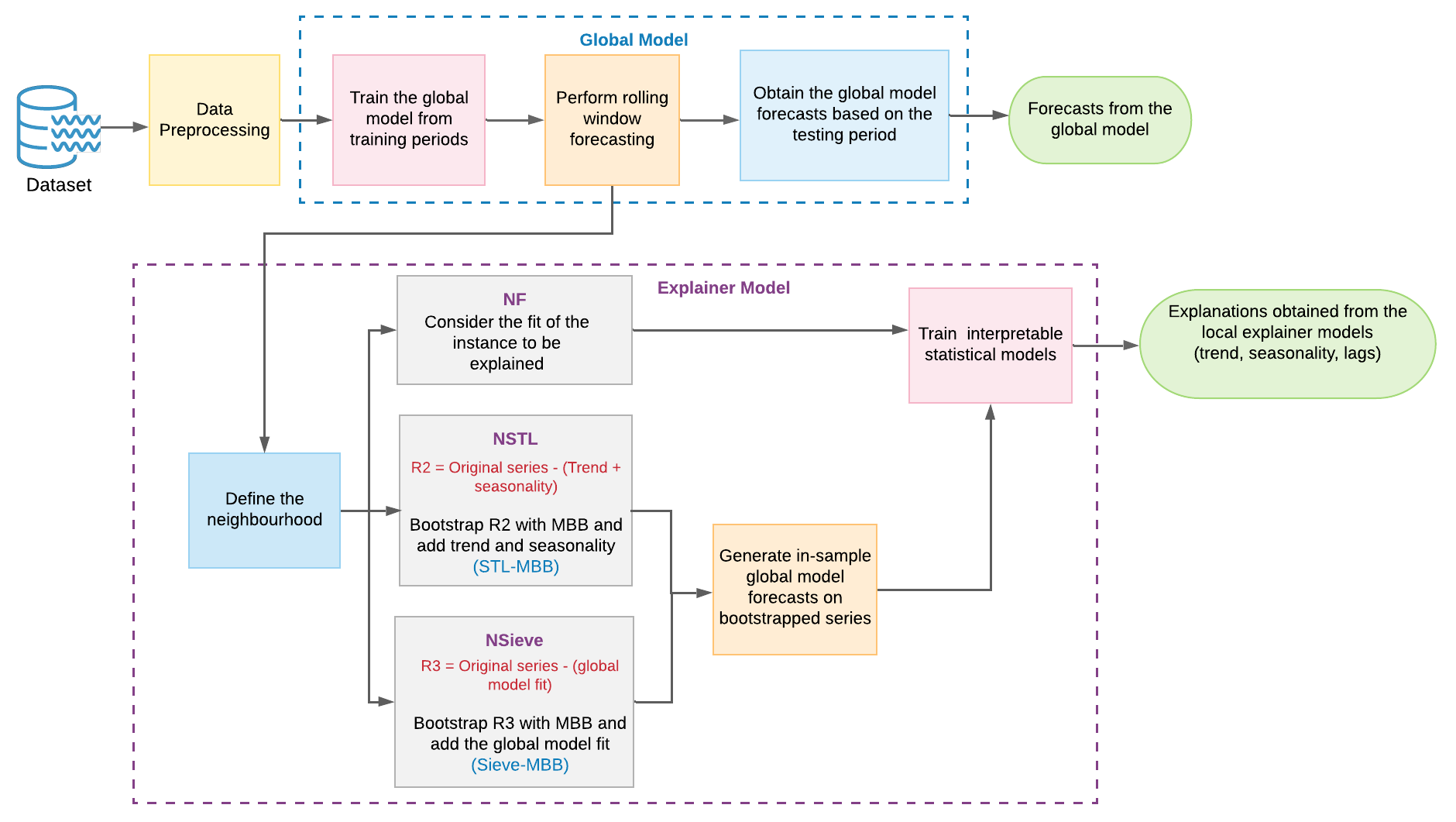}
	\caption{An overview diagram of the framework to generate local explanations for the global model predictions.}
	\label{Fig:framework}
\end{figure*}

Transferring the local interpretability concept to the forecasting domain, as the background model we want to use a GFM $f$ that has been trained on a set of time series $Y=\{y_1,\ldots,y_p\}$, where $p$ is the total amount of time series available. Now, let us consider the predictions $\hat{y}$ of a particular time series $y$, obtained from the GFM as $f(y) = \hat{y}$. We want to build a local explainer $e$ as one or more traditional forecasting models, such as ETS, ARIMA, THETA, such that $e(y)\approx \hat{y}$.
Thus, it is straightforward to use the original time series $y$ as the neighbourhood, as the local explainer $e$ is able to be fitted on a single time series. We then need to find a sampling procedure from the neighbourhood. 

Let us first consider the case where $f$ is a purely autoregressive model that has no internal state and only uses lagged values as inputs, such as a feedforward neural network. Then, we are in a similar situation to the regression case discussed in Section~\ref{sec:locint}. We can now first assume a procedure that does not sample any data, and all data that we use is data from the training set that is within the neighbourhood, i.e., $\tilde{X}$ and $X_{\textit{filt}}$ are equal. In the forecasting context, this means that the target values we use to fit $e$ are the predictions of the GFM along the time series, i.e., the in-sample fit of $f$ on $y$. As we want to use time series models as local explainers, we choose also the input for $e$ to be the in-sample fit (instead of the original series).
Thus, a first approach is to simply fit $e$ on the in-sample fit of $f$ on $y$. We use in this work the one-step-ahead in-sample fit of $f$, even if $f$ produces multi-step forecasts.
The Figure~\ref{Fig:one_step_ahead_forecast} illustrates the one-step-ahead in-sample fit of the global model.
Here, each red dot indicates a one-step-ahead forecast or the fit of the global model. 
Note that this approach is easily extendable to GFMs that are not purely autoregressive.

This approach of not sampling data has the potential drawback that the surrogate model cannot assess the generalisation capabilities of the global model, which can be a problem especially if the global model is overfitting. In extreme cases, depending on the GFM, the fit can not bear any meaningful information, e.g., it can be identical to the original time series if the GFM produces a perfect in-sample fit.
However, motivated by the findings of \cite{montero2020principles} that GFMs typically lead to worse fitting than per-series models, but generalise better, in this paper we explore at first this simple solution without sampling. In the remainder of the paper, we call this the Neighbourhood-Fit (NF) method.
Then, we propose two methods based on bootstrapping, as discussed in the next section, thus, defining three types of explainer methods in total, based on their neighbourhood generation procedures. An overview diagram of the proposed framework is shown in Figure~\ref{Fig:framework}. 



\subsection{Neighbourhood generation with Bootstrapping}

Bootstrapping time series has the difficulty that the series can be non-stationary and have autocorrelation. Non-stationarity can be addressed by time series decomposition as proposed in the work of \citet{bergmeir2016bagging}, or using sieve bootstrapping procedures \citep{buhlmann1997sieve}, e.g., as done by \citet{cordeiro2009forecasting}. 
The autocorrelation in the series can be addressed with the Moving Block Bootstrap \citep[MBB,][]{kunsch1989jackknife}, or again with the sieve bootstrap, fitting a model and then bootstrapping the residuals, assuming that they are uncorrelated.
Also, more sophisticated bootstrapping procedures have been introduced for time series, such as the tapered block bootstrap~\citep{paparoditis2001tapered}, the extended tapered block bootstrap~\citep{shao2010extended}, the dependent wild bootstrap \citep[DWB,][]{shao2010dependent}, or the linear process bootstrap \citep[LPB,][]{mcmurry2010banded}. 
However, in this work, we focus on the common and established approaches based on MBB, decomposition, and sieve bootstrapping.

In particular, we use the procedure from \citet{bergmeir2016bagging} which uses the MBB technique together with STL decomposition. Here, a remainder is extracted after performing the STL decomposition on the original series. Then, the decomposed trend and seasonality are added to the bootstrapped series. 
We call this approach Neighbourhood STL-MBB (NSTL) method in the remainder of the paper.  
Another possibility we use is a sieve bootstrapping procedure where we fit a model and then bootstrap the residuals.
Similar to the NSTL approach, the MBB technique is used to generate the new set of remainders.
We call this approach Neighbourhood Sieve-MBB (NSieve) in the following.
For NSTL and NSieve we use the \texttt{bld.mbb.bootstrap} and \texttt{MBB} functions, respectively, from the \texttt{forecast} package in R \citep{hyndman2018forecast,Khandakar2008Automatic}.
%
After generating the neighbourhood as bootstrapped series, we generate in-sample global model forecasts on these series, and fit our local explainer models to these. 
Note that, when considering the local explainer methods using bootstrapping procedures, one local explainer model is built per bootstrapped time series, so that it is necessary to select local explainers where the explanations can be summarized across the bootstrapped series.

\subsection{Global model forecasting procedure}

Many different types of global forecasting models have been proposed recently, from earlier works of \cite{smyl2016data}, \cite{wen2017multi}, \cite{salinas2020deepar}, and \cite{bandara2020forecasting}, to the winning method of the M4 by \cite{smyl2020hybrid}, to \cite{oreshkin2019n} and others. As our work is agnostic towards the global model used, we aim for a relatively standard Long Short-Term Memory (LSTM) network implementation, following the guidelines of \cite{hewamalage2021recurrent}. We detail the exact global model that we use in the following.



\subsubsection{Recurrent Neural Networks with Long Short-Term Memory Cells}\label{subsubsec: RNN-LSTM}
RNNs are neural networks that have a feedback loop to propagate an inner state~\citep{elman1990finding}, which makes them appropriate for sequential modeling tasks such as Natural Language Processing or speech recognition. LSTMs are a special type of RNNs that solve problems of vanishing and exploding gradients in traditional RNNs by gating mechanisms, which enable them, together with the feedback loops, to capture non-linear long-term temporal dependencies in sequences.
They have recently gained popularity in the forecasting space in particular as the winning method of the M4 competition~\citep{smyl2020hybrid} is based on LSTMs.
Following the guidelines of \citet{hewamalage2021recurrent}, we use a stacked architecture combined with LSTMs with peephole connections~\citep{gers2000learning}, and perform a number of preprocessing steps.

%

\subsubsection{Data Preprocessing}

Training and validation periods are chosen according to \citet{hewamalage2021recurrent}. 
We reserve a separate section from each time series as the validation period to perform hyper-parameter tuning. 
The global LSTM model is built across a group of time series and may contain observations in different value ranges. 
Consequently, it may be beneficial to perform a normalisation to scale the observations.
We perform a mean scaling transformation as the normalisation strategy, which divides the whole time series $y_i$ by its corresponding mean value as indicated in Equation~\ref{eq:mean normalise}.

\begin{equation}\label{eq:mean normalise}
     y_{i,normalised} = \frac{y_i}{\frac{1}{k}\sum_{t=1}^{k}y_{i,t}} 
\end{equation}

Thereafter, we take the logarithm of each series to stabilise the variance in the time series. 
As the time series in our experiments are all non-negative and may contain zero values, we apply the log transformation as defined in Equation~\ref{eq:log_scale}.

\begin{equation}\label{eq:log_scale}
    y_{i,log\_scaled} =
    \begin{cases}
            \log(y_i), & \text{if min$(Y) > 0$}, \\
            \log(y_i+1), &  \text{if min$(Y) = 0$}
    \end{cases}
\end{equation}
Here, $Y$ is the whole set of time series and  $y_{i,log\_scaled}$ is the time series $i$ after logarithmic scaling.


%

In our approach we use Fourier terms as additional inputs~\citep{hyndman2018forecasting}, where a set of sine and cosine terms are used to model the seasonality of the time series~\citep{harvey199310}.  
Following~\citet{bandara2020lstm} we train the RNN model together with the pre-processed time series and Fourier terms, where the Fourier terms are added as an external variable to capture the periodic effects of the time series.


Assume $y_{i,t}$ is an observation of time series $y_i$ in a particular point of time $t$. 
Equation~\ref{eq:fourier_term} defines the approach of approximating the periodic terms for the observation $y_{i,t}$. 

\begin{equation}\label{eq:fourier_term}
\sin{\left(\frac{2\pi kt}{s1}\right)},\cos{\left(\frac{2\pi kt}{s1}\right)},...,\sin{\left(\frac{2\pi kt}{s_n}\right)},\cos{\left(\frac{2\pi kt}{s_n}\right)}
\end{equation}

Here, $n$ is the number of seasonalities in the time series, such that $s_n$ represents the $n$th seasonality of time series $y_i$, and $k = 1,\ldots, K$, so that there is a total amount of $K$ sine and cosine pairs for each seasonality. 
The variable $K$ controls the regularity of the periodic patterns, and we determine its value using a grid-search algorithm within the range of 1 to 25.

We use the Multi-Input Multi-Output (MIMO) strategy when producing forecasts.
Here, we transform a time series $y_i$ into multiple pairs of $(T_i-n-m)$ records, where $T_i$ is the length of series $y_i$, $n$ is the length of the input window, and $m$ is the length of the output window.
Thus, each record has an amount of $(m + n)$ observations.
The MIMO strategy produces forecasts for the whole output window at once. Hence, it avoids the accumulation of error in each forecasting step of the time series when training neural networks. 
Following the aforementioned procedure, to train the global model, we use $(T_i-m)$ observations from time series $y_i$ and reserve the last output window as the validation period for parameter tuning.

\subsubsection{Global model training}

The RNN models we use have several hyperparameters that need to be tuned, namely: cell dimension, mini-batch size, maximum number of epochs, epoch size, number of hidden layers, L2-regularisation weight, standard deviation of random normal initialiser, and standard deviation of the Gaussian noise.
In line with~\citet{hewamalage2021recurrent}, we perform automatic hyperparameter tuning using Sequential Model based Algorithm Configuration optimisation~\citep[SMAC,][]{hutter2011sequential}.
We use COntinuous COin Betting (COCOB) as the learning algorithm to select an optimal learning rate when training the global model~\citep{orabona2017training}.  
The model generates the final forecasts after training the model using the optimal set of hyperparameters.
To address parameter uncertainty due to the randomly chosen initial weights and the stochastic nature of the optimisation procedure of the NN, we train on 10 different Tensorflow graph seeds and average the resulting forecasts that are then used as final forecasts in the evaluation, effectively building an ensemble of 10 models. Though also the hyperparameter tuning has uncertainty associated, we do not re-tune the hyperparameters for separate seeds as it would be computationally too expensive.  
%

\subsubsection{Post processing global model forecasts}


The post-processing of the forecasts is as follows. First, we take the exponential of the values to reverse the log transformation.
Then, we subtract 1 if the original dataset contains zero values. 
Afterwards, we de-normalise by multiplying the forecasts of every series by its corresponding mean.
Then, if the original dataset that we consider contains integer values, we round the forecasts to the nearest integer, assuming that integer forecasts are required in these use cases.
Finally, if the original dataset that we consider contains only non-negative values, we convert all negative values to zero. 
%


\subsection{Local explainer models}

In this section, we detail the forecasting models that we use as surrogate explainer models.
The main focus of our work is not to argue that these models are explainable, but our work allows us to lift GFMs to the same level of explainability as these simpler models. We assume that the models discussed in the following are interpretable through components such as relevant lags, trend, seasonality, and others.


\subsubsection{Exponential Smoothing}

Exponential Smoothing is a very popular family of univariate forecasting models. It uses decompositions and weighted averages of past observations, where the weights are decaying exponentially over time~\citep{brown1961fundamental}.
In its simplest form, called simple exponential smoothing, it doesn't model trend or seasonality. More sophisticated models have been introduced subsequently to model trends and seasonalities~\citep{brown1957exponential,winters1960forecasting} in additive and multiplicative ways.

In this work, we use the \texttt{ets} function from the \texttt{forecast} package~\citep{hyndman2018forecast,Khandakar2008Automatic} in the R programming language.
The function fits different exponential smoothing models with different trends and seasonalities, and then choose one model based on a criterion such as AICc. Thus, both the type of model (for example, a model with a linear trend and a multiplicative seasonality) as well as the decomposed time series into trend, seasonality, and remainder, can be used to interpret the predictions.

\subsubsection{STL Decomposition}
Seasonal and Trend decomposition using Loess~\citep[STL,][]{cleveland1435502stl} is a statistical approach of decomposing a time series $x_t$ additively into three components of seasonality $S_t$, trend $T_t$, and residuals $R_t$, as shown in Equation~\ref{eq:STL}.

\begin{equation}\label{eq:STL}
    x_t = S_t + T_t + R_t
\end{equation}


The components can be used to interpret the forecast. 
STL is limited to series with a single seasonality. If a series has multiple seasonalities, multiple STL (MSTL) decomposition can be used.

Similar to STL, MSTL also decomposes the time series into seasonal components, trend, and remainder. It extracts multiple seasonal components by applying the STL procedure iteratively. 
Thus, Equation~\ref{eq:STL} can be extended to Equation~\ref{eq:MSTL}. 

\begin{equation}\label{eq:MSTL}
    x_t = S^1_t + S^2_t +...+ S^n_t + T_t + R_t
\end{equation}

Here, $S^1_t,\ S^2_t,\ ...,\ S^n_t$ denote the multiple seasonal components.
Again, we consider the decomposed components of the time series as interpretable features of a forecast. 
We use the implementations of STL and MSTL available in the functions \texttt{stl} and \texttt{mstl} from the \texttt{stats} and \texttt{forecast} packages in R~\citep{team2019,hyndman2018forecast,Khandakar2008Automatic}.

\subsection{Linear Autoregressive models}

We also consider univariate linear autoregressive (AR) models since they can be fitted both locally and globally (also called pooled regression, PR), but with the same interpretability through their coefficients.
In the pooled regression, coefficients are the same across all time series. 
In our experiments we use PR as a local explainer in the NSTL and NSieve bootstrapping explainer models. In this context, the term ``global'' means training across the bootstrapped series of the instance that needs to be explained. Hence, it produces explanations that are still local to the time series in consideration. 
To implement the PR models, we use the~\texttt{glm} function in the~\texttt{stats} package of the R programming language \citep{team2019}.

\subsubsection{Dynamic Harmonic Regression ARIMA}

Dynamic Harmonic Regression (DHR) is a statistical forecasting approach which is often used to forecast series with long seasonal periods such as weekly and daily series, where ARIMA and ETS tend to produce less accurate results~\citep{hyndman2018forecasting}.
In the approach of DHR-ARIMA, we use Fourier terms to model the seasonality and ARIMA errors to model the short-term time series dynamics while assuming that the seasonal pattern is constant over time.
In terms of interpretation, the value of K Fourier terms, selected through grid search, gives an indication of the complexity of the seasonal pattern. The value of the first 2 coefficients, i.e., for cos(2$\pi$t/s) and sin(2$\pi$t/s), can be interpreted as the relative importance of the corresponding seasonal period.
We use the \texttt{auto.arima} function from the \texttt{forecast} package in R~\citep{hyndman2018forecast,Khandakar2008Automatic} together with the Fourier terms generated by the function \texttt{fourier} from the same package to implement the model.

\subsubsection{Trigonometric Box-Cox ARMA Trend Seasonal Model: TBATS}
TBATS is a state-of-the-art approach to forecast time series with complex multiple seasonalities~\citep{delivera2011forecasting}.
To model complex multiple seasonalities, TBATS uses trigonometric representation terms based on Fourier series. 
Additionally, TBATS has the ability to model both linear and non-linear time series with single seasonalities, multiple seasonalities, high-period seasonalities, non-integer seasonalities and dual calender effects. 
As a local explainer, TBATS performs a decomposition of the series into complex seasonal patterns, trend, and irregular components. 
We use TBATS in its implementation in the \texttt{tbats} function from the \texttt{forecast} package in R.

\subsubsection{PROPHET}
PROPHET~\citep{Taylor2017-ck} is another popular univariate forecasting software based on decompositions of the time series. 
It fits an additive decomposition model where non-linear trends are fitted with multiple seasonalities (e.g., yearly, weekly, and daily seasonality), together with holiday effects, as defined in Equation~\ref{eq:prophet}. 

\begin{equation}\label{eq:prophet}
    x_t = S^1_t + S^2_t +...+ S^n_t + T_t + R_t + H_t
\end{equation}

Here, $H_t$ represents the holiday effects.
Most importantly, in PROPHET, human interpretable parameters (i.e., decomposed seasonalities, trend, holiday effects) can be encompassed based on domain knowledge to improve the forecast.
In this paper, we use the implementation of PROPHET in the \texttt{prophet} package in R~\citep{Taylor2017-ck}.

\subsubsection{THETA}
The THETA method was initially introduced by~\citet{assimakopoulos2000theta} and attracted the interest of forecasting practitioners after winning the M3 forecasting competition~\citep{makridakis2000m3}. It was later shown by~\cite{hyndman2003unmasking} that the method is in fact identical to simple exponential smoothing with drift, where the drift parameter is chosen as half the slope of a linear trend that is fitted to the data. As the method is simple and therewith interpretable, but yet has been very successful in the past, we use it as a local explainer.
We use in our work the implementation of THETA in the \texttt{thetaf} function from the \texttt{forecast} package in R. 
 



\section{Experimental Setup}
\label{sec:Experiments}

In this section we describe the benchmark datasets, benchmark local models, error metrics, and the evaluation criteria of the proposed LoMEF framework used for our experiments. 

\subsection{Datasets}
\label{sec:Datasets}

The aim of our paper is to explain predictions of global models in situations where the global models have high accuracy and are preferred over traditional univariate models.
Therefore, from the literature~\citep{hewamalage2021recurrent,bandara2020forecasting} and preliminary experiments not reported here, in this work we selected five benchmark time series datasets on which the global forecasting models perform well. 
A brief description of the benchmark datasets is as follows. 


\begin{itemize}
    \item Ausgrid-Energy half-hourly dataset (AusGridHH): This dataset consists of data related to the energy consumption of 300 households in Australia~\citep{AusGrid2019-dl}. 
    In our experiments, we select data related to the general consumption (GC letter code in the dataset), which is one of the primary energy consumption categories.  
    Here, we extract a subset of half-hourly data from 2012 July to 2012 October. 
    \item Ausgrid-Energy weekly dataset (AusGridW): This dataset is an aggregated version of the Ausgrid-Energy half-hourly dataset. Before aggregation, we omitted one series which has missing values for more than eight consecutive months. Thus, this dataset contains 299 weekly series. For this dataset, we use the full period for which data is available, namely from 2010 July to 2013 June.
    \item San Francisco Traffic hourly dataset (SFTrafficH): This dataset contains the hourly traffic data on San Francisco Traffic Bay area freeways from 2015 to 2016~\citep{Caltrans2020-dr}.
    \item NN5 weekly dataset (NN5W): This dataset is an aggregated version of the NN5-daily dataset used in the NN5 forecasting competition which is data on cash withdrawals from ATMs in the UK~\citep{taieb2012review}.
    The missing values of the original daily dataset on a particular day are replaced by the median across the same days of the week along the whole series.
    Then the daily time series are aggregated into weekly.
    \item Kaggle Web Traffic daily dataset (WebTrafficD): This dataset was used in the Kaggle Wikipedia Web Traffic forecasting competition~\citep{Google2017-cc}. It consists of daily web traffic (number of hits) of around 145,000 Wikipedia pages.
    Following the configurations used in the work of~\citet{hewamalage2021recurrent}, in our experiments, we use the first 997 series from the dataset, for the period of 1st July 2015 to 31st December 2016.
    Differing from the other datasets in our study, this dataset is count data. Thus, we round the final forecasts to the closest non-negative integer. Furthermore, as the dataset doesn't exhibit seasonality, in this dataset we don't use Fourier terms with the RNN.
    \item Kaggle Web Traffic100 daily dataset (WebTraffic100D): For the experiments of NSTL and NSieve methods, randomly sampled 100 records were selected from WebTrafficD dataset, since it consumes considerable amount of memory when dealing with the full dataset.

\end{itemize}

Table~\ref{tab:summary_datasets} shows summary statistics of the datasets used in this study. We note that all series have equal lengths across a dataset.
Similar to the heuristics proposed by~\citet{hewamalage2021recurrent,bandara2020forecasting}, we choose the input window size as $1.5\times\mathrm{horizon}$, assuming that the horizon has been chosen in relation to the autocorrelation structure and seasonality of the series. 


\begin{table}[!htpb]
\centering
\caption{Dataset Statistics} 
\label{tab:summary_datasets}
\scalebox{0.75}{
\begin{tabular}{lcccc}
  \hline
Dataset &  No of series & Length  & Seasonal Cycles & Horizon \\ 
  \hline
NN5W & 111  & 105	&	52&	8\\ 
AusGridW & 299	&148 &	52	&8\\ 
WebTrafficD & 997 &	549 &	no&	60\\ 
SFTrafficH & 963&	726 &	(24, 168)&	24\\  
AusGridHH & 300&	4704&(48, 336)&	96\\ 
   \hline
\end{tabular}
}
\end{table}

\subsection{Benchmark models}

We use ETS, TBATS, DHR-ARIMA, PROPHET, STL-ETS, THETA, and MSTL-ETS as the benchmark models.
Table~\ref{tab:benchmark_models} illustrates the benchmark models used for each dataset. 
Note that the benchmark models (local models) are trained on the actual data, in contrast with the same models used as local explainers, trained on the respective sampled data for explanation.
We select the local models and the local explainer models for each dataset considering the length of the series, number of seasonal periods, and type of the seasonality.
For example, seasonal series use seasonal local models and seasonal local explainers, and non-seasonal series use non-seasonal local models and non-seasonal local explainer models, accordingly.
In our experimental setup, for the non-seasonal datasets like the WebTrafficD dataset, we use non-seasonal ETS and THETA as the local models and the local explainers. 
To forecast the time series with complex seasonalities we use TBATS, PROPHET, and DHR\_ARIMA in the NN5W, AusGridW, SFTrafficH, and AusGridHH datasets. 
Moreover, we categorise the benchmark models based on the number of seasonalities in the dataset. 
We use MSTL\_ETS for hourly datasets with multiple seasonalities and we used STL\_ETS for weekly datasets with a single seasonality.

\begin{table}[!htpb]
	\centering
	\caption{Benchmark Models for the Datasets} 
	\label{tab:benchmark_models}
	\scalebox{0.85}{
		\begin{tabular}{ll}
			\hline
			Dataset  & Benchmark Models \\ 
			\hline
			NN5W &	TBATS, DHR\_ARIMA, PROPHET, PR \\ 
			AusGridW &	TBATS, DHR\_ARIMA, PROPHET, STL\_ETS\\ 
			WebTrafficD	& ETS, THETA, PR\\ 
			SFTrafficH &	TBATS, PROPHET, DHR\_ARIMA, MSTL\_ETS\\  
			AusGridHH &TBATS, PROPHET, DHR\_ARIMA, MSTL\_ETS \\ 
			\hline
		\end{tabular}
	}
\end{table}

\subsection{Error Metrics}
\label{subsec:error_metrics}

To measure the performance of our models, we use three different error metrics commonly used in the forecasting space, namely the mean absolute scaled error \citep[MASE, ][]{hyndman2006another}, Root Mean Squared Error (RMSE), and Mean Absolute Error (MAE). 
The MASE is a standard scale-free error measure common in forecasting, used in, e.g., the M4~\citep{smyl2020hybrid} competition. 
The MAE and RMSE are scaled measures that therewith avoid many of the problems that scale-free measures have, and we use them as our datasets have the same scale across different series. 
Equations~\ref{eq:mase}, \ref{eq:rmse}, and \ref{eq:mae} define the MASE, RMSE, and MAE measures, respectively~\citep{hyndman2006another}. 

\begin{equation}\label{eq:mase}
\text{MASE} = \dfrac{1}{h}{\dfrac{\sum_{t=M+1}^{M+h}|\hat{y}_t - y_t|}{\dfrac{1}{M-S}\sum_{t=S+1}^{M}|y_t - y_{t-S}|}}
\end{equation}

\begin{equation}\label{eq:rmse}
    \text{RMSE} = \sqrt{\dfrac{\sum_{t=1}^{h}(\hat{y}_t - y_t)^2}{h}}
\end{equation}

\begin{equation}\label{eq:mae}
    \text{MAE} = \dfrac{\sum_{t=1}^{h}|\hat{y}_t - y_t|}{h}
\end{equation}

Here, $y_t$ and $\hat{y}_t$ denote the observation and the forecast at time $t$, respectively. 
$M$ is the number of data points in the training series, $S$ is the seasonality of the dataset, $h$ is the forecast horizon,

%
%

%

\subsection{Evaluating the local explainers}

%

When it comes to evaluating the local explainers, a question is what are the desirable qualities of such explainers. Following the general literature on local explainability \citep{yang2019bim,carvalho2019machine}, we evaluate our local explainers along the following dimensions.
 
\begin{itemize}
	\item Fidelity - How well does the explainer approximate the forecast of the GFM?
	\item Accuracy - How well does an explainer forecast unseen data?
	\item Stability - How consistent are the explanations of the explainer across multiple runs? 
	\item Comprehensibility - How interpretable are the explanations?
\end{itemize}


To perform the evaluations, we first calculate error measures among the actual observations, global model forecasts, local statistical benchmark model forecasts, and the local explainer model forecasts, as follows. Here, $Error(y_t,\hat{y}_t)$ can be one of MAE, RMSE, or MASE. 

\begin{itemize}
    \item $Error(global,\ explainer)$: Error between the global model forecasts and forecasts generated from the local explainer model. 
    \item $Error(actual,\ global)$: Error between the actual observations and the forecasts generated from the global model. 
    \item $Error(actual,\ local)$: Error between the actual observations and the forecasts generated from the local statistical model. 
    \item $Error(global,\ local)$: Error between the global model forecasts and the forecasts generated from the local statistical model. 
    \item $Error(actual,\ explainer)$: Error between the actual observations and the forecasts generated from the local explainer model. 
\end{itemize}

Based on these primary error measures, we can then calculate secondary error measures as follows.

\begin{equation}\label{eq:fidelity_1}
    Fidelity\_Actual = Error(global,explainer) - Error(global,actual)
\end{equation}
 $Fidelity\_Actual$ as defined in Equation~\ref{eq:fidelity_1} verifies whether the local explainer forecasts are closer to the GFM forecasts than the GFM forecasts to the actuals. 
Thus, a value of $Fidelity\_Actual$ of less than zero shows that the explainer model resembles the behaviour of the global model on the given time series better than the global model is able to predict the actuals.

\begin{equation}\label{eq:fidelity_2}
    Fidelity\_Local = Error(global,explainer) - Error(global,local)
\end{equation}
$Fidelity\_Local$ as defined in Equation~\ref{eq:fidelity_2} assesses whether the local explainer forecasts are closer (indicated by a negative value of the measure) to the GFM forecasts than the local model forecasts to the GFM forecasts. This can be seen as a sanity check that the explainer actually resembles the GFM, compared with a similar model trained directly on the data without involvement of the GFM.


\begin{equation}\label{eq:closer_model_to_exp}
Fidelity\_with\_Explainer = Error(global,explainer) - Error(local,explainer)
\end{equation}
Equation~\ref{eq:closer_model_to_exp} shows the formula to verify whether the local explainer is closer to the global model than the local model. If the value is less than zero, the explainer model is closer to the global model than the local model and vise versa. 
The value of $Fidelity\_with\_Explainer$ being negative means that the explainer models perform well for the considered dataset when explaining the forecast of the global model. 
The $Fidelity\_with\_Explainer$ measure evaluates the fidelity of the explainer by examining whether the explainer is mimicking the global model better than the local model.  

For local explainers, fidelity is an important property, and the different fidelity measures enable us to evaluate different aspects of the fidelity of the local explainer. Next, we define measures of accuracy for the local explainers. Accuracy is arguably less important as the main goal of a local explainer is to explain well, not to predict well.

\begin{equation}\label{eq:global_local}
Acc\_Global\_LocalModel = Error(actual,global) - Error(actual,local)
\end{equation}
Equation~\ref{eq:global_local} shows the formula to verify whether the global model performs better than the local model. If the value is less than zero, the global model performs better than the local model and vise versa.
This measure does not directly evaluate the local explainer, but for a complete picture and to put the other evaluations into context, it is necessary to know whether the global model outperforms the local models.

\begin{equation}\label{eq:exp_local}
    Acc\_Explainer\_LocalModel = Error(actual,explainer) - Error(actual,local)
\end{equation}

$Acc\_Explainer\_LocalModel$ defined in Equation~\ref{eq:exp_local} assesses whether the local explainer performs better than the local model when forecasting. If the value is less than zero, the explainer model performs better than the local model and vise versa. 
Thus, $Acc\_Explainer\_LocalModel$ evaluates the accuracy of the explainer compared to the local model. 

\begin{equation}\label{eq:exp_global}
    Acc\_Explainer\_GlobalModel = Error(actual,explainer) - Error(actual,global)
\end{equation}
Equation~\ref{eq:exp_global} shows the formula to assess whether the local explainer performs better than the global model. If the value is less than zero, the explainer model performs better than the global model and vise versa.
In other words, $Acc\_Explainer\_GlobalModel$ evaluates the accuracy of the explainer compared to the global model. 

To understand the broader view of the model performance, we further calculate the mean and median values across time series of the per-series performance measures ($Fidelity\_Actual, \\ Fidelity\_Local,Fidelity\_with\_Explainer, Acc\_Global\_LocalModel, Acc\_Explainer\_LocalModel, \\Acc\_Explainer\_GlobalModel$ ). 
These measures evaluate the fidelity and accuracy dimensions of the local explainers. 

To evaluate the stability of the local explainer models, we calculate $Error(global, explainer)$ over independent runs of the NSTL and NSieve bootstrapping explainer methods, as well as the NF method, and afterwards measure the stability of these errors with boxplots and the interquartile range (IQR) over these runs. We note that, as the NF method executes the local explainer only on a single time series, here stability effectively measures the stability of the local forecasting model.

To evaluate the comprehensibility of the explainer models we follow a functionally grounded evaluation procedure (proxy tasks) in our experiments \citep{doshi2017towards}.
This technique can be applied when the class of local explainer models has already been validated as being interpretable, e.g., using a user study or as in our case assuming that these models are well-established in practice for interpretations.  
In particular, our main aim is to argue that with the proposed methodology, global models can be made as interpretable as traditional local statistical models. Arguing that traditional statistical models are interpretable is not the main focus of our paper, though we show many examples of how these models can be used for interpretations in practice, e.g., with components such as trend, seasonality, model coefficients and other model characteristics. Those are considered as the proxy for the quality of the explanations, and a qualitative evaluation with examples is given in Section~\ref{subsec: examples}.

%

\section{Results and Discussion}
\label{sec:results}

In this section, first, we discuss the quantitative results of our study, measuring the quality of the local explainers in terms of the fidelity, accuracy, and stability dimensions of the explainer models. 
Then, we provide examples of explanations of the local explainers for a qualitative assessment to examine the quality of local explainer models in terms of their comprehensibility.

\begin{table}[htpb]
	\vspace{-2cm}
	\centering
	\caption{Mean and median performance measures based on the RMSE error measure for datasets on different local explainer models. Asterisks indicate statistically significant results, boldface font indicates negative values (which are desirable).} 
	\label{tab:mm_rmse}
	\scalebox{0.7}{
		\begin{tabular}{llcccccc}
			\hline
			Dataset 
			&\multicolumn{1}{p{1.5cm}}{\centering  Local\\ Explainer} 
			& Fidelity\_Actual 
			& Fidelity\_Local 
			& \multicolumn{1}{p{2.0cm}}{\centering Fidelity\_with\\\_Explainer}
			&\multicolumn{1}{p{2.0cm}}{\centering Acc\_Global\_\\LocalModel} 
			&\multicolumn{1}{p{2.2cm}}{\centering Acc\_Explainer and \\ Local Model} 
			& \multicolumn{1}{p{2.2cm}}{\centering Acc\_Explainer and \\Global Model}\\ 
			\hline
			\multicolumn{8}{c}{Mean Values}\\
			\hline
			NN5W& PROPHET & \textbf{-9.154}\anote& \textbf{-3.152}\anote & \textbf{-1.908} &\textbf{ -0.739} &\textbf{ -0.222} & 0.517 \\ 
			& DHR\_ARIMA & \textbf{-10.353}\anote & \textbf{-3.311}\anote & \textbf{-2.233}\anote & \textbf{-0.422} & \textbf{-0.450} & \textbf{-0.028} \\ 
			& TBATS & -\textbf{11.119}\anote & \textbf{-4.206}\anote & \textbf{-1.787} & \textbf{-3.041}\anote & \textbf{-2.111}\anote & 0.931 \\ 
			AusGridW& PROPHET &\textbf{ -9.170}\anote & \textbf{-2.950}\anote & 8.039 & \textbf{-4.352} & \textbf{-2.209}\anote & 2.143 \\ 
			& DHR\_ARIMA & \textbf{-11.223}\anote & \textbf{-3.658}\anote & 5.289 & \textbf{-3.088} & \textbf{-1.689} & 1.399 \\ 
			& TBATS & \textbf{-12.602}\anote & \textbf{-2.231} & 2.104 & \textbf{-3.248} &\textbf{ -0.741} & 2.508 \\ 
			& STL\_ETS & \textbf{-9.967}\anote & \textbf{-5.769}\anote & 0.678 &\textbf{ -6.435}\anote & \textbf{-4.240}\anote & 2.194 \\ 
			WebTrafficD & ETS & \textbf{-20.205} & \textbf{-9.778}\anote & \textbf{-8.909}\anote & \textbf{-4.706} & \textbf{-4.708} & \textbf{-0.002} \\ 
			& THETA &\textbf{ -20.322}\anote & \textbf{-6.033}\anote & \textbf{-5.365}\anote & \textbf{-2.248} &\textbf{ -2.198} & 0.050 \\ 
			SFTrafficH & TBATS & \textbf{-0.007}\anote & \textbf{-0.008}\anote & \textbf{-0.006}\anote & \textbf{-0.003}\anote & \textbf{-0.004}\anote &\textbf{ -0.001}\anote \\ 
			& PROPHET &\textbf{ -0.006}\anote & \textbf{-0.007}\anote & \textbf{-0.004}\anote & \textbf{-0.008}\anote & \textbf{-0.006}\anote & 0.002 \\ 
			& DHR\_ARIMA & \textbf{-0.006}\anote & \textbf{-0.009}\anote & \textbf{-0.005}\anote & \textbf{-0.005}\anote & \textbf{-0.007}\anote &\textbf{ -0.001}\anote \\ 
			& MSTL\_ETS & \textbf{-0.002} & \textbf{-0.006}\anote & \textbf{-0.003}\anote & \textbf{-0.006}\anote & \textbf{-0.004}\anote & 0.002 \\ 
			AusGridHH & TBATS & \textbf{-0.136}\anote & \textbf{-0.156}\anote & \textbf{-0.133}\anote & \textbf{-0.088}\anote & \textbf{-0.090}\anote & \textbf{-0.002} \\ 
			& PROPHET & \textbf{-0.106}\anote & \textbf{-0.178}\anote & \textbf{-0.136}\anote & \textbf{-0.120}\anote & \textbf{-0.110}\anote & 0.010 \\ 
			& DHR\_ARIMA & \textbf{-0.119}\anote & \textbf{-0.079}\anote &\textbf{ -0.026}\anote & \textbf{-0.029}\anote & \textbf{-0.029}\anote & 0.000 \\ 
			& MSTL\_ETS & 0.013 & \textbf{-0.135}\anote & \textbf{-0.169}\anote & \textbf{-0.186}\anote & \textbf{-0.085} & 0.101 \\ 
			\hline
			\multicolumn{8}{c}{Median Values}\\
			\hline
			NN5W& PROPHET &\textbf{ -7.885}\anote & \textbf{-2.204}\anote & \textbf{-1.592} & \textbf{-1.031} &\textbf{ -0.586} & \textbf{-0.020} \\ 
			& DHR\_ARIMA & \textbf{-8.673}\anote &\textbf{ -2.494}\anote & \textbf{-1.658}\anote & \textbf{-0.650} & \textbf{-0.272} & \textbf{-0.070} \\ 
			& TBATS &\textbf{ -9.474}\anote &\textbf{ -3.405}\anote & \textbf{-0.916} & \textbf{-1.615}\anote &\textbf{ -0.936}\anote & 0.824 \\ 
			AusGridW& PROPHET & \textbf{-6.447}\anote & \textbf{-1.819}\anote & 4.226 & \textbf{-2.344 }& \textbf{-1.382}\anote & 0.920 \\ 
			& DHR\_ARIMA & \textbf{-7.451}\anote &\textbf{ -2.891}\anote & 2.017 &\textbf{ -1.544} & \textbf{-1.382} & 0.480 \\ 
			& TBATS &\textbf{ -7.229}\anote & \textbf{-2.233} & 1.454 & \textbf{-1.767} & \textbf{-0.672} & 1.010 \\ 
			& STL\_ETS &\textbf{ -6.696}\anote & \textbf{-4.741}\anote & \textbf{-1.678} & \textbf{-4.298}\anote & \textbf{-3.191}\anote & 1.160 \\ 
			WebTrafficD & ETS &\textbf{ -9.534} & \textbf{-2.028}\anote & \textbf{-2.000}\anote & \textbf{-0.112} & \textbf{-0.083} & 0.000 \\ 
			& THETA & \textbf{-9.567}\anote &\textbf{ -1.757}\anote & \textbf{-1.648}\anote & 0.000 & 0.000 & 0.000 \\ 
			SFTrafficH & TBATS & \textbf{-0.005}\anote & \textbf{-0.006}\anote & \textbf{-0.004}\anote & \textbf{-0.001}\anote & \textbf{-0.002}\anote & \textbf{-0.002}\anote \\ 
			& PROPHET & \textbf{-0.004}\anote & \textbf{-0.006}\anote &\textbf{ -0.003}\anote & \textbf{-0.007}\anote & \textbf{-0.006}\anote & 0.001 \\ 
			& DHR\_ARIMA & \textbf{-0.004}\anote & \textbf{-0.008}\anote &\textbf{ -0.003}\anote & \textbf{-0.004}\anote & \textbf{-0.005}\anote & \textbf{-0.002}\anote \\ 
			& MSTL\_ETS & \textbf{-0.001} &\textbf{ -0.006}\anote & \textbf{-0.003}\anote & \textbf{-0.005}\anote & \textbf{-0.004}\anote & 0.000 \\ 
			AusGridHH & TBATS & \textbf{-0.117}\anote &\textbf{ -0.092}\anote & \textbf{-0.071}\anote &\textbf{ -0.032}\anote &\textbf{ -0.037}\anote & \textbf{-0.002} \\ 
			& PROPHET & \textbf{-0.091}\anote &\textbf{ -0.134}\anote & \textbf{-0.100 }\anote &\textbf{ -0.082}\anote & \textbf{-0.069}\anote & 0.004 \\ 
			& DHR\_ARIMA & \textbf{-0.101}\anote &\textbf{ -0.061}\anote & \textbf{-0.021}\anote & \textbf{-0.013}\anote &\textbf{ -0.014}\anote & \textbf{-0.002} \\ 
			& MSTL\_ETS & \textbf{-0.067} & \textbf{-0.139}\anote & \textbf{-0.116}\anote &\textbf{ -0.116}\anote &\textbf{ -0.093} & 0.008 \\ 
			\hline
		\end{tabular}
	}
\end{table}



\begin{table}[htpb]
	\centering
	\caption{Mean and median performance measures based on RMSE error measures of the datasets on different local explainer models using NSTL approach. Asterisks indicate statistically significant results, boldface font indicates negative values (which are desirable).} 
	\label{tab:mm_rmse_bm1}
	\scalebox{0.75}{
		\begin{tabular}{llcccccc}
			\hline
			Dataset 
			&\multicolumn{1}{p{1.5cm}}{\centering  Local\\ Explainer} 
			& Fidelity\_Actual 
			& Fidelity\_Local 
			& \multicolumn{1}{p{2.0cm}}{\centering Fidelity\_with\\\_Explainer}
			&\multicolumn{1}{p{2.0cm}}{\centering Acc\_Global\_\\LocalModel} 
			&\multicolumn{1}{p{2.2cm}}{\centering Acc\_Explainer and \\ Local Model} 
			& \multicolumn{1}{p{2.2cm}}{\centering Acc\_Explainer and \\Global Model}\\ 
			\hline
			\multicolumn{8}{c}{Mean Values}\\
			\hline
			NN5W & PR & \textbf{-6.755}\anote & 0.967 & \textbf{-0.266} & \textbf{-2.365} & \textbf{-1.865} & 0.501 \\ 
				& ETS & - & - & -  & - &-&- \\ 
			WebTraffic100D & PR & \textbf{-14.156}\anote & \textbf{-1.547}\anote &\textbf{ -1.661}\anote & \textbf{-0.399} & \textbf{-0.416} & \textbf{-0.017} \\ 
			& ETS & \textbf{-15.282}\anote &\textbf{ -10.474}\anote & \textbf{-10.391}\anote & \textbf{-4.773} & \textbf{-4.824} & \textbf{-0.050} \\ 
			
			\hline
			\multicolumn{8}{c}{Median Values}\\ 
			\hline
			NN5W & PR &\textbf{ -6.443}\anote & \textbf{-0.079} & 0.918 &\textbf{ -0.946} & \textbf{-0.626} & 0.115 \\ 
				& ETS & - &- & - & - &-&- \\
			WebTraffic100D 	& PR &\textbf{ -9.302}\anote & \textbf{-0.753}\anote &\textbf{ -0.957}\anote & 0.052 & 0.075 & 0.000 \\ 
			& ETS &\textbf{ -10.186}\anote & -\textbf{2.946}\anote & \textbf{-3.027}\anote & \textbf{-0.018} &\textbf{ -0.030} & 0.000 \\ 
		
			\hline
		\end{tabular}
	}
\end{table}


\begin{table}[htpb]
	\centering
	\caption{Mean and median performance measures based on RMSE error measures of the datasets on different local explainer models using NSieve approach. Asterisks indicate statistically significant results, boldface font indicates negative values (which are desirable).} 
	\label{tab:mm_rmse_bm2}
	\scalebox{0.75}{
		\begin{tabular}{llcccccc}
			\hline
			Dataset 
			&\multicolumn{1}{p{1.5cm}}{\centering  Local\\ Explainer} 
			& Fidelity\_Actual 
			& Fidelity\_Local 
			& \multicolumn{1}{p{2.0cm}}{\centering Fidelity\_with\\\_Explainer}
			&\multicolumn{1}{p{2.0cm}}{\centering Acc\_Global\_\\LocalModel} 
			&\multicolumn{1}{p{2.2cm}}{\centering Acc\_Explainer and \\ Local Model} 
			& \multicolumn{1}{p{2.2cm}}{\centering Acc\_Explainer and \\Global Model}\\ 
			\hline
			\multicolumn{8}{c}{Mean Values}\\
			\hline
			NN5W & PR & \textbf{-7.072}\anote & 0.650 & 3.090 &\textbf{ -2.365 }& \textbf{-0.044} & 2.321 \\
			& ETS & - &- & - & - &-&- \\ 
			WebTraffic100D & PR & \textbf{-12.744}\anote & \textbf{-0.167} &\textbf{ -1.505} & \textbf{-0.405} & 0.289 & 0.694 \\
			& ETS &\textbf{ -12.929}\anote & \textbf{-8.121} & \textbf{-10.508}\anote & \textbf{-4.773} &\textbf{ -3.845} & 0.929 \\ 
			 
			\hline
			\multicolumn{8}{c}{Median Values}\\ 
			\hline
			NN5W & PR & \textbf{-6.049}\anote & 0.821 & 2.555 & \textbf{-0.946} & 0.226 & 1.371 \\ 
			& ETS & - &- & - & - &-&- \\
			WebTraffic100D & PR &\textbf{ -7.877}\anote & \textbf{-0.394} & \textbf{-0.854} & 0.061 & 0.292 & 0.149 \\ 
			& ETS &\textbf{ -9.438}\anote & \textbf{-1.286} & \textbf{-3.000} & \textbf{-0.018}\anote & 0.000 & 0.360 \\ 
			
			\hline
		\end{tabular}\textbf{}
	}
\end{table}


\subsection{Quantitative results}\label{subsec:quantitative results}

In the following, we discuss the quantitative results of the NF, NSTL and NSieve explainer methods. 

\subsubsection{Evaluating accuracy and fidelity of the explainer methods}

\sloppy{
	Table~\ref{tab:mm_rmse} shows the mean and median values of the performance measures $Fidelity\_Actual, Fidelity\_Local,Fidelity\_with\_Explainer, Fidelity\_with\_Explainer$, $Acc\_Explainer\_GlobalModel$, $Acc\_Explainer\_LocalModel$, and  $Acc\_Global\_LocalModel$ relevant to the NF explainer method for the RMSE error measure.
For the results of the explainer methods based on the bootstrapping procedures, Table~\ref{tab:mm_rmse_bm1} shows the mean and median values of the performance measures relevant to the NSTL explainer method and Table~\ref{tab:mm_rmse_bm2} shows the mean and median values of the performance measures relevant to the NSieve explainer method for the RMSE error measure. Note that in the latter two tables, no results are reported for the NN5W dataset with ETS explainer, as ETS cannot be applied to weekly datasets due to their long seasonal cycles. In the following, we discuss the results related to RMSE.
The results for the MASE and MAE error measures are similar and lead to the same conclusions. They are reported in Tables~\ref{tab:mm_mase}, \ref{tab:mm_mae}, \ref{tab:mm_mase_bm1}, \ref{tab:mm_mae_bm1}, \ref{tab:mm_mase_bm2} and \ref{tab:mm_mae_bm2} in the Appendix.

	Considering the NF explainer method, most of the mean and the median $Fidelity\_Actual$ values of the RMSE error measure are negative except the mean $Fidelity\_Actual$ value of the MSTL local explainer which is used in the AusGridHH dataset. 
	With regards to NSTL and NSieve explainer methods, both mean and median $Fidelity\_Actual$ measures show negative values. 
	This implies that in most cases the forecasts produced by the global models are closer to the forecasts produced by the explainers than the actual values. 
	Therefore,the analysis of $Fidelity\_Actual$ concludes that the explainers approximate the global model considerably better than the actuals.
}

In all three explainer methods for both mean and median, the $Fidelity\_Local$ measures show negative values for the NN5W dataset which uses the bootstrapping procedures in the explainer methods, which assures that the local model forecasts approximate the forecasts of the global model better than the local model forecasts approximate the global model under all three techniques.
In other words, the explainer model approximates the global model better than the local model approximates the global model. Therefore, when $Fidelity\_Local$ is negative, the explainer model performs better than the local model.

The mean and median $Fidelity\_with\_Explainer$ measures show mostly negative values except for few cases in the NF explainer method in the AusGridHH dataset, the NSTL method in WebTraffic100D, and the NSieve explainer methods in the NN5W and WebTraffic100D datasets.  
This implies that the forecasts from the explainers are considerably closer to the forecasts of the global model compared with the forecasts of the local models. 
In other words, the explainer model approximates the global model better than the local model.
Therefore, when $Fidelity\_with\_Explainer$ is negative, the explainer model mimics the global model well.

The mean and median $Acc\_GlobalLocalModel$ measures show negative values in all three explainer methods across all the explainer models which implies that the actual values are closer to the global model forecasts than the local model forecasts.
This indicates that the global model is better than the local model, in terms of accuracy measured against the actual values (which is not surprising, as the datasets have been selected in a way that global models perform well, as outlined in Section \ref{sec:Datasets}). 

The mean and median $Acc\_Explainer\_LocalModel$ performance measures show negative values for all three types of error measures with regards to all three explainer methods except for the mean MASE error measure in the NN5W dataset on the NSieve explainer method which implies that the forecasts of the explainer model are closer to the actual values than the forecasts of the local model. Thus, the performance of the explainer model is better than the local model when compared with the actual values, which is an interesting property and an interesting avenue to achieve more accurate but still simple local models. 

Considering all three types of explainer methods, the mean and median $Acc\_Explainer\_GlobalModel$ measures show mostly positive values except for few cases which implies that the forecasts of the global model are closer to the actual values than the forecasts of the explainer model. 
Thus, the performance of the global model is better than the explainer model, which is not surprising as the main aim of an explainer model is to explain well, not to achieve the best possible accuracy.

We furthermore perform a Student's t-Test as implemeted in the function \texttt{t.test} in the~\texttt{stats} package of the R programming language \citep{team2019} to evaluate the statistical significance of our results. In particular, we perform one-sided tests to test if values are significantly smaller than zero. Assuming independence in the results between the datasets and local explainers, but dependence across error measures (RMSE, MASE, MAE), we use a Bonferroni correction separately on each error measure to control the family-wise error across multiple tests, and to avoid spurious results. We use an initial significance level of 0.05, and as we perform 300 tests in total per error measure (RMSE, MASE, MAE), this leads to a corrected significance level of $\alpha \approx 1.6\times 10^{-4}$. The statistically significant results are marked with an asterisk (*) in Tables~\ref{tab:mm_rmse},~\ref{tab:mm_rmse_bm1} and ~\ref{tab:mm_rmse_bm2} of the main paper and Tables~\ref{tab:mm_mase},~\ref{tab:mm_mase_bm1},~\ref{tab:mm_mase_bm2},~\ref{tab:mm_mae},~\ref{tab:mm_mae_bm1} and~\ref{tab:mm_mae_bm2} in the Appendix. 
Furthermore, in the Appendix, Tables~\ref{tab:mean_rmse_stat}, \ref{tab:mean_mase_stat}, \ref{tab:mean_mae_stat}, \ref{tab:mean_rmse_stat_bm1}, \ref{tab:mean_mase_stat_bm1}, \ref{tab:mean_mae_stat_bm1}, \ref{tab:mean_rmse_stat_bm2}, \ref{tab:mean_mase_stat_bm2}, and \ref{tab:mean_mae_stat_bm2} show the corresponding unadjusted $p$-values. In these tables with $p$-values, values in a boldface font indicate values that are statistically significant. 

In summary, by analysing $Fidelity\_Actual$, $Fidelity\_Local$, and $Fidelity\_with\_Explainer$, we conclude that the local explainer model is explaining the global model for the given benchmark datasets and their particular local explainers well.
By analysing the results of $Acc\_Explainer\_LocalModel$, we have shown that there is a possibility to increase the accuracy of the forecasts of the local models by training them as local explainers following our methodology.

\subsubsection{Evaluating the stability of the explainer methods}

\begin{figure}[h]
	\centering
	\subfloat[PR local explainer on NN5W dataset]{
		\includegraphics[width=0.45\textwidth]{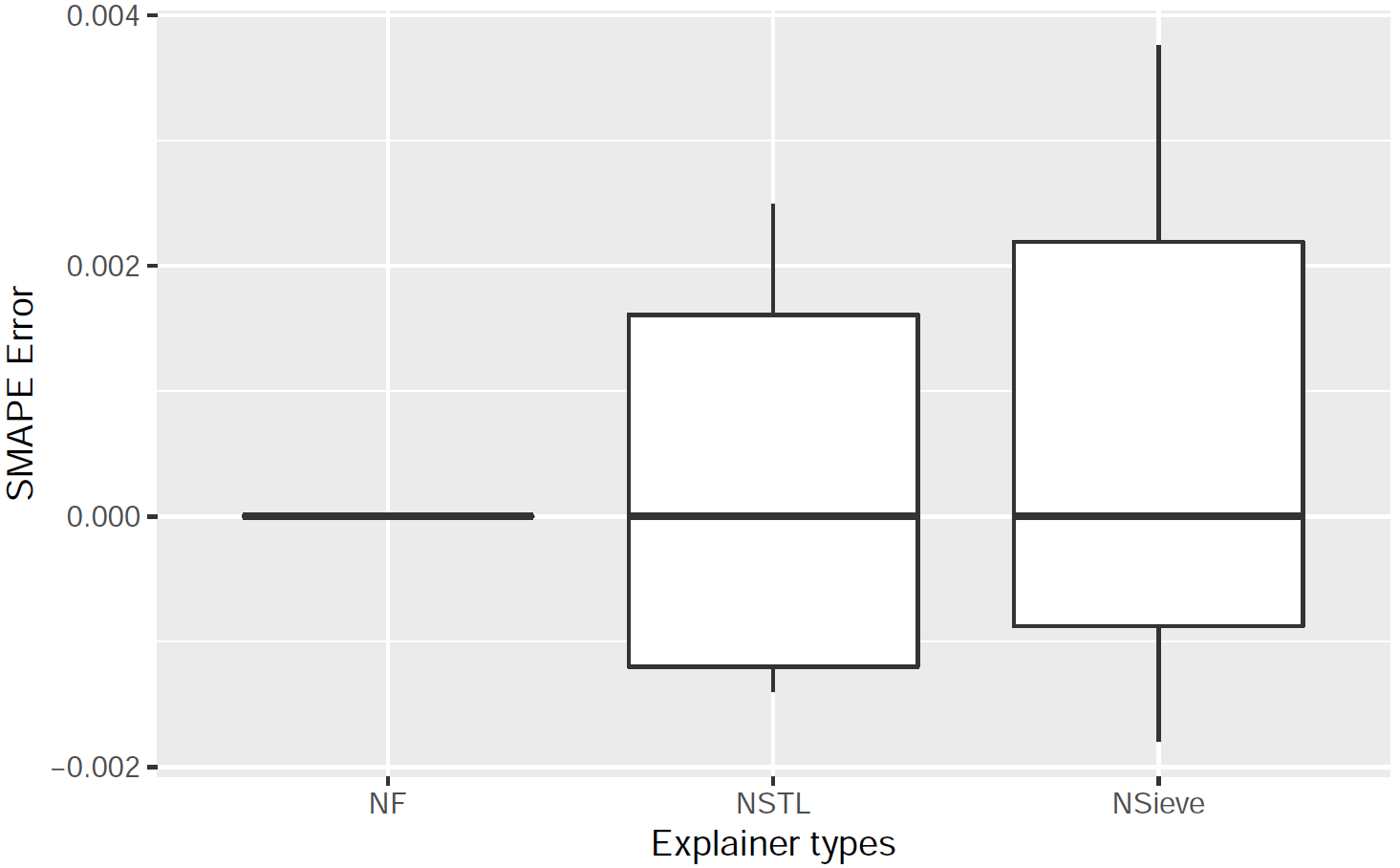}
		\label{fig:PR_NN5W_stability}}
	\subfloat[PR local explainer on WebTraffic100D dataset]{
		\includegraphics[width=0.45\textwidth]{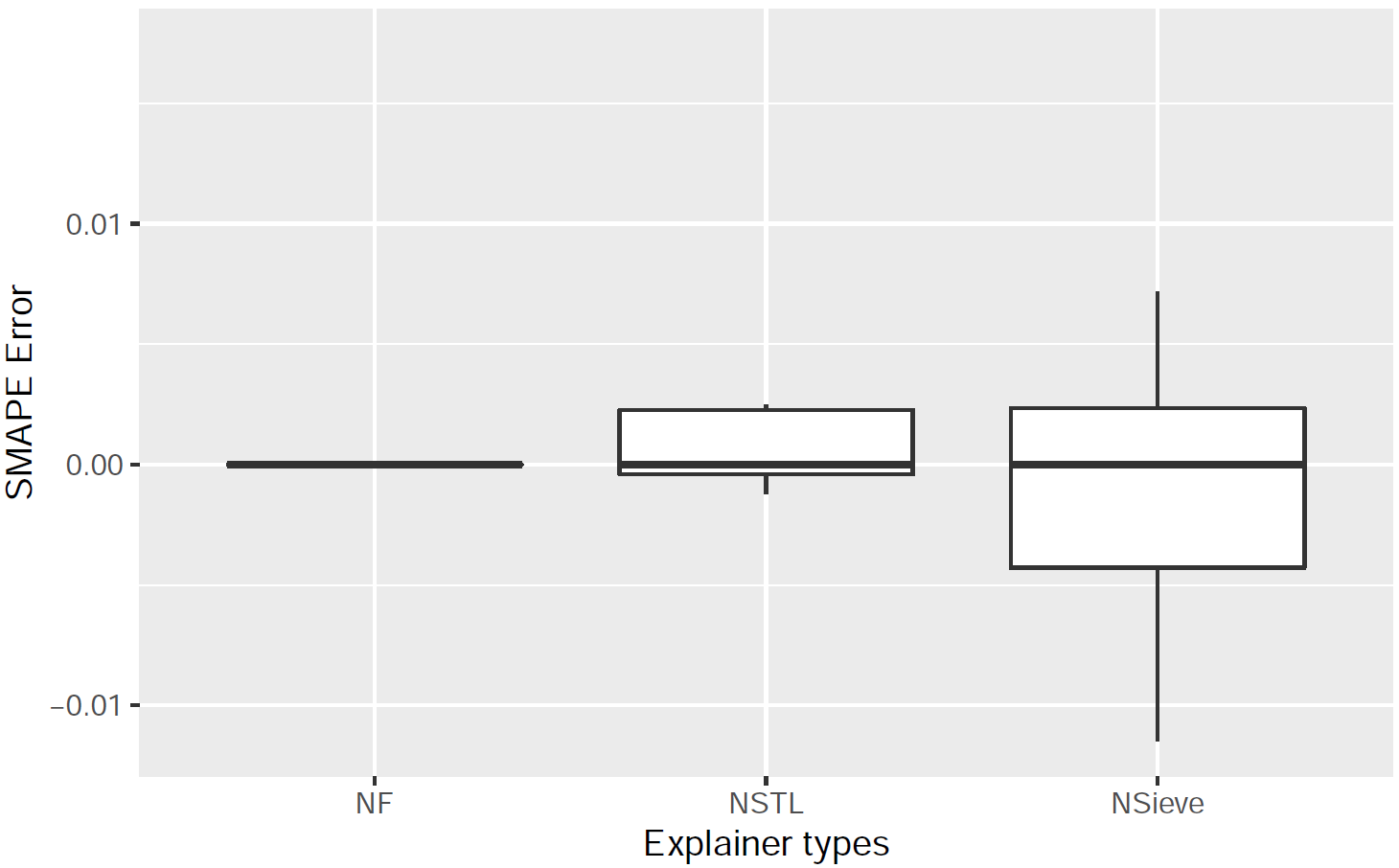}
		\label{fig:PR_WebTraffic100D_stability}}
	\qquad
	\subfloat[ETS local explainer on WebTraffic100D dataset]{
		\includegraphics[width=0.45\textwidth]{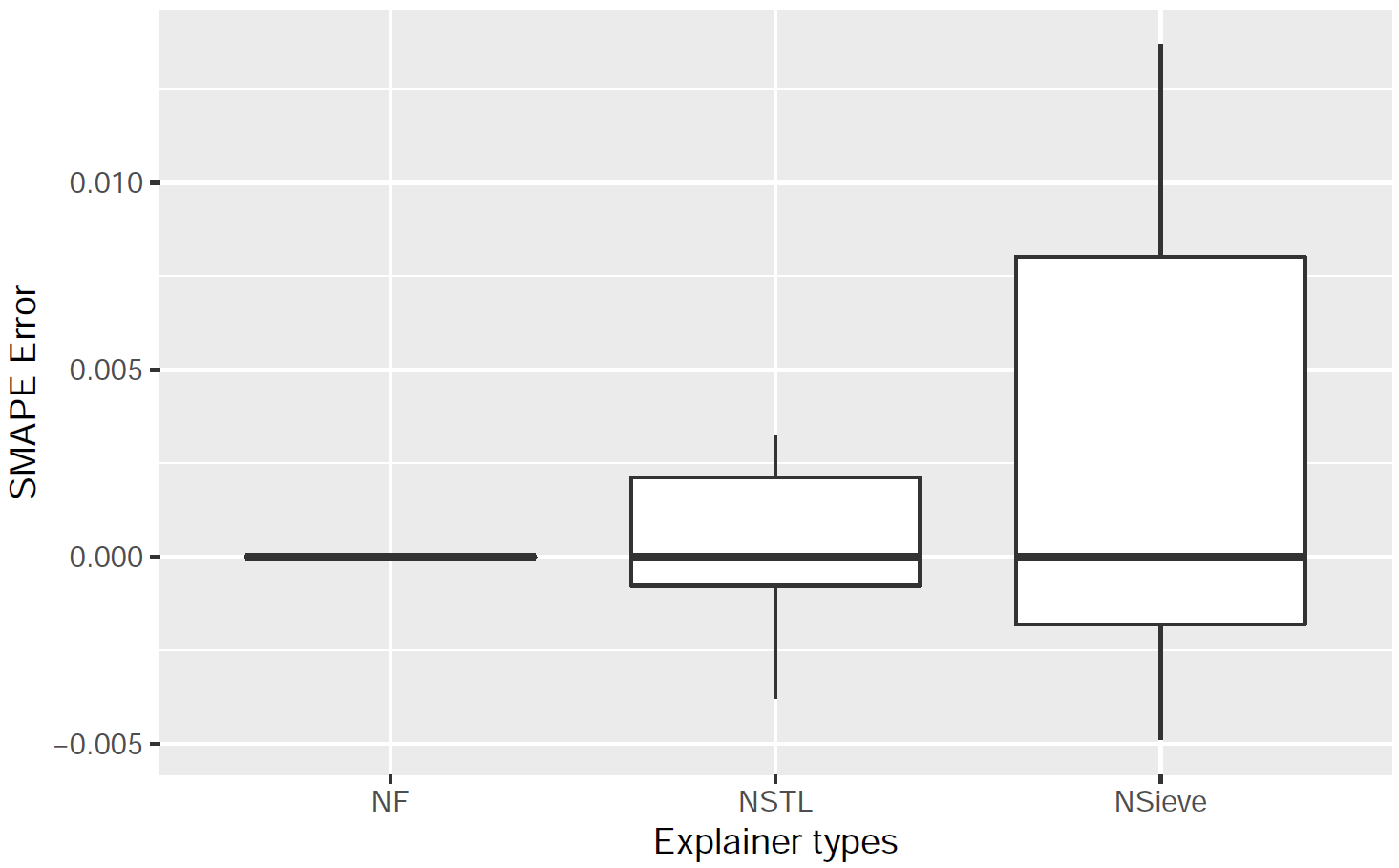}
		\label{fig:ETS_WebTraffic100D_stability}}
	\caption{Distribution of RMSE error across the independent runs of proposed explainer methods}
	\label{fig:consistency_plot}
\end{figure}

For this experiment, we consider the same datasets and the explainer models that were used when producing explanations using the NF, NSTL and NSieve methods. 
Figure~\ref{fig:consistency_plot} shows boxplots of the median RMSE between the global model forecast and the PR local explainer on the NN5W dataset, the PR local explainer on the WebTraffic100D dataset, and the ETS local explainer on the WebTraffic100D dataset, respectively, over 10 independent runs. Note that for NF, the PR model is fitted on a single time series and effectively falls back to an AR model.
 Table~\ref{tab:stability} shows the according IQRs. From the plots and the table, we see that the NSTL approach in general is more stable than the NSieve approach, and the NF approach is unsurprisingly the most stable approach as it has no uncertainty introduced through a bootstrapping procedure, and only reflects the uncertainty in the fitting of the local model. 

%

\begin{table}[htpb]
	\centering
	\caption{Stability of proposed explainer methods based on bootstrapping } 
	\label{tab:stability}
	\scalebox{0.75}{
		\begin{tabular}{llrrr}
			\hline
			Dataset & Local Explainer & NF & NSTL & NSieve \\ 
			\hline
			NN5W & PR & 0.000 & 0.229 & 0.212 \\ 
			& ETS& - & - & - \\ 
			WebTraffic100D & PR & 0.000 & 0.073 & 0.187 \\ 
			& ETS& 0.000 & 0.069 & 0.209 \\ 
			\hline
		\end{tabular}
	}
\end{table}

%


\subsection{Qualitative results}\label{subsec: examples}


This section evaluates the comprehensibility of the explanations qualitatively with example explanations generated using the different local explainers for all three explainer methods. 


\subsubsection{Example explanations for the NF method}


In the following, we discuss four different examples for different local explainer models that can explain the forecasts via decompositions.

\paragraph{Example 1: Decomposition explanations using a TBATS local explainer}


\begin{figure}[!htb]
	\centering
	\subfloat[TBATS local explainer forecasts]{
		\includegraphics[width=0.8\textwidth]{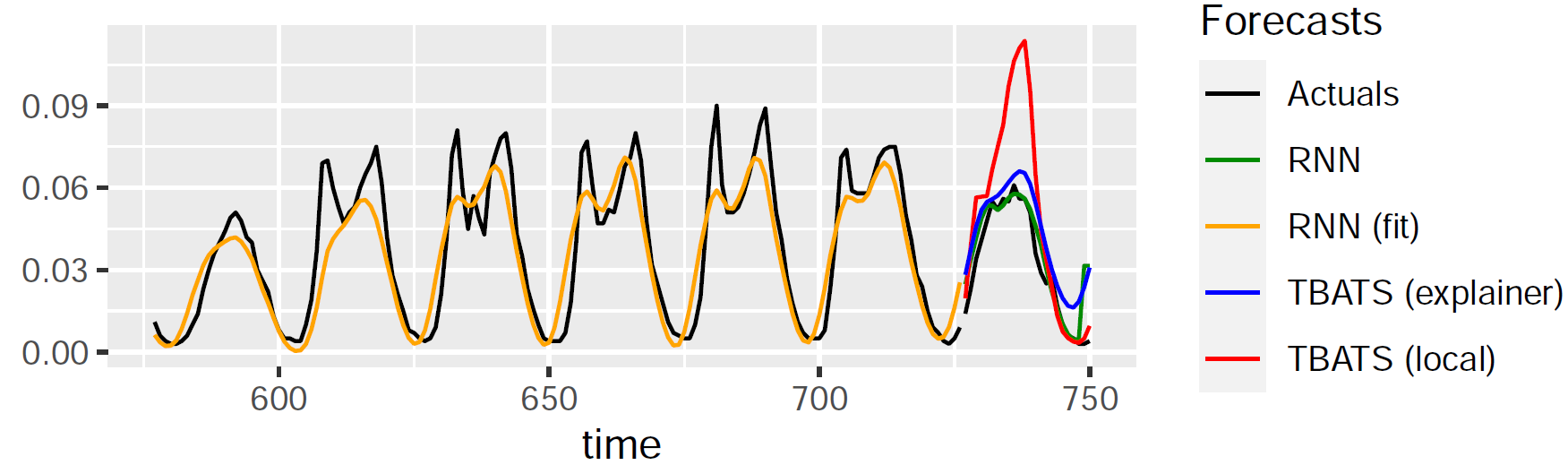}
	\label{Fig:Tbats_forecasts}}
		\qquad
	\subfloat[Decomposition plot for TBATS interpretation]{
			\includegraphics[width=0.6\textwidth]{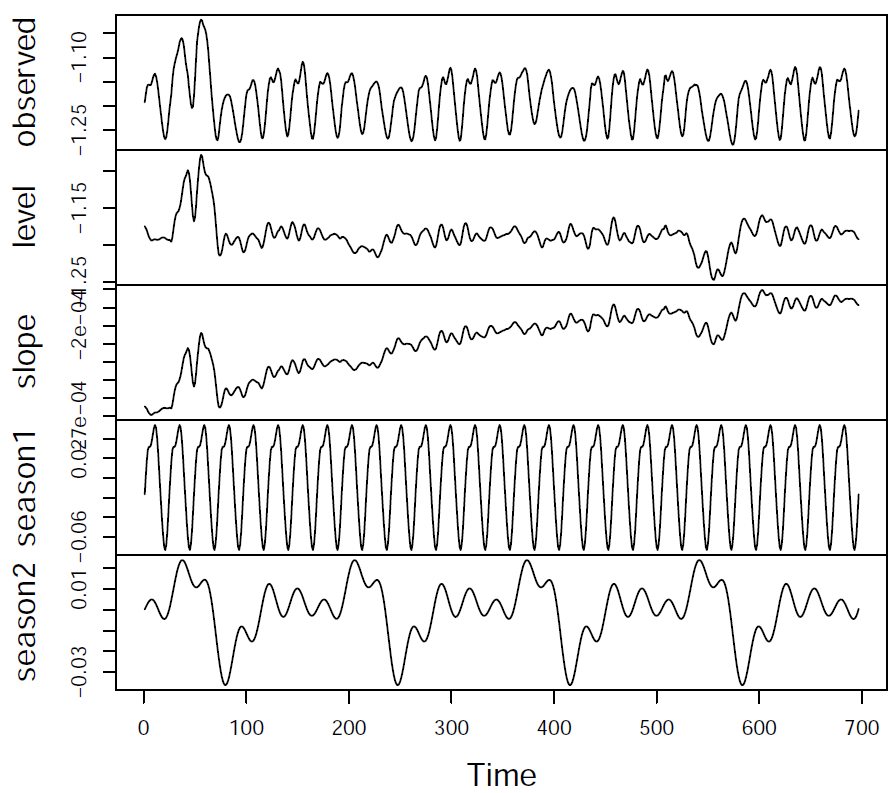}
	\label{Fig:Tbats_decomposition}}
	\caption{TBATS local explainer forecasts and interpretation.}
\end{figure}


Figure~\ref{Fig:Tbats_forecasts} shows a plot of forecasts based on the TBATS model for a single time series of the San Francisco Traffic hourly dataset.  It allows us to visually verify fidelity and accuracy as follows. We can see that the global RNN model outperforms the local TBATS model since the RNN forecasts are closer to the actuals than the local TBATS model.
Moreover, we can observe that the TBATS explainer model approximates the global model better than the TBATS local model. 
Both of these observations show that the TBATS explainer is performing as expected.
Figure~\ref{Fig:Tbats_decomposition} gives an example of explanations of the global model forecasts provided by the TBATS explainer model. It provides components (ideally of the forecasts but in our plot of the training set), namely daily (season2) seasonalities, weekly (season1) seasonalities, level, and slope, to give insights into the forecast.

\paragraph{Example 2: Decomposition explanations using MSTL local explainer}

\begin{figure}[!htb]
	\centering
	\subfloat[MSTL local explainer forecasts]{
		\includegraphics[width=0.8\textwidth]{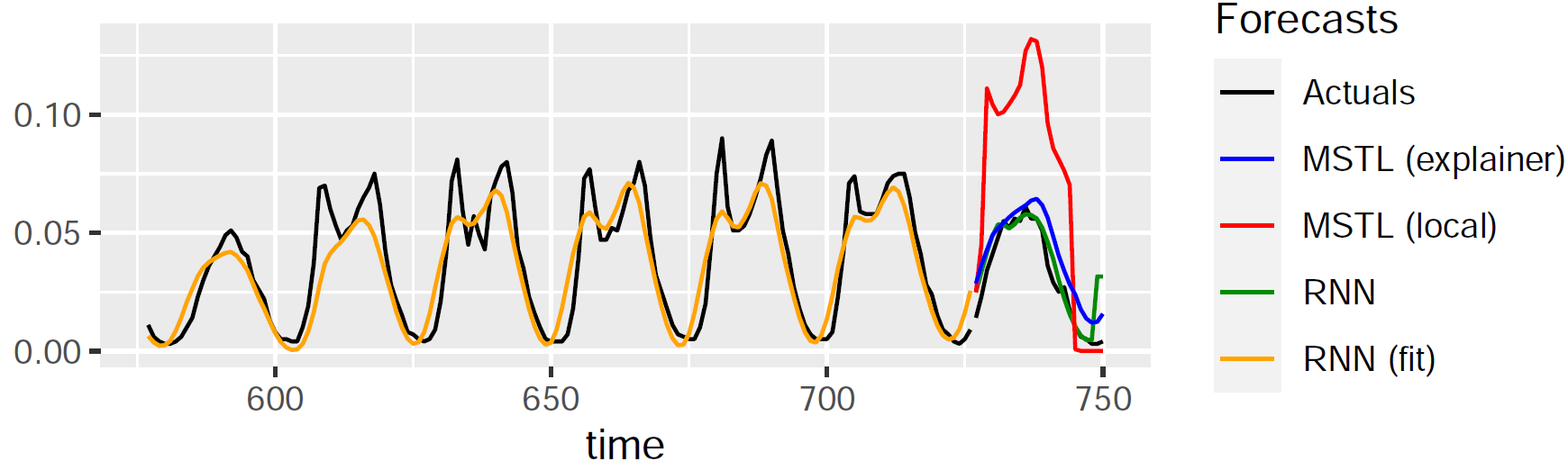}
	\label{Fig:mstl_forecasts}}
		\qquad
	\subfloat[Decomposition plot for MSTL interpretation]{
			\includegraphics[width=0.6\textwidth]{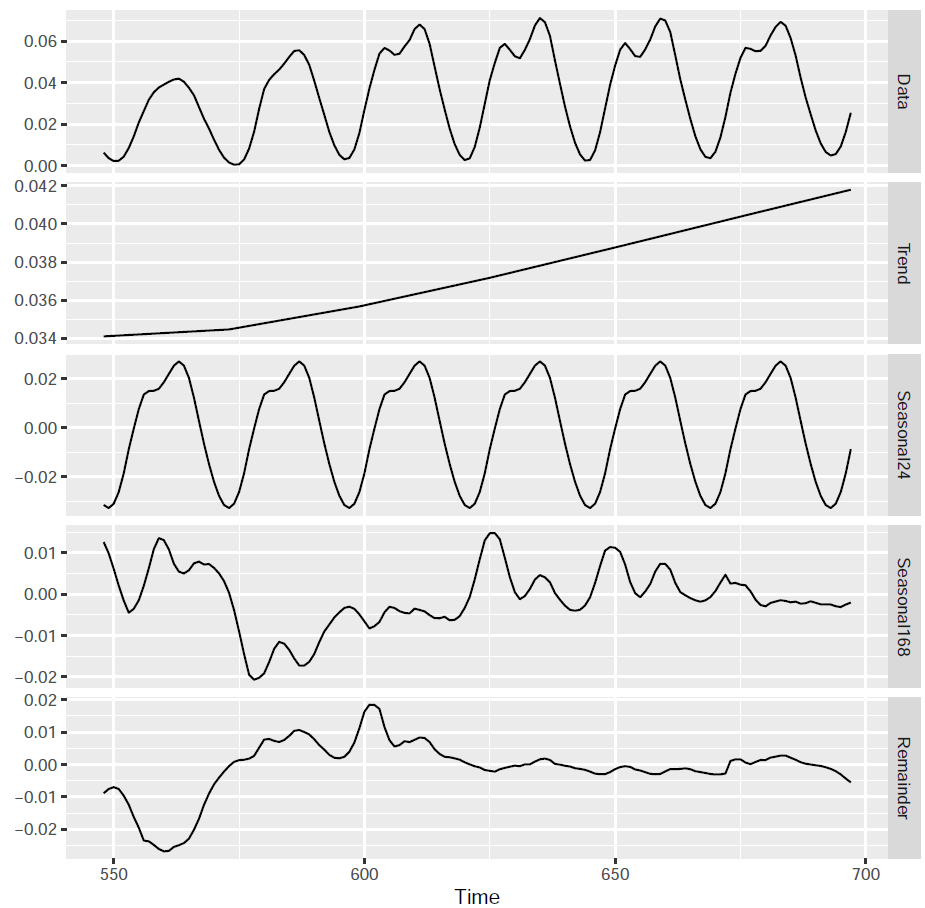}
	\label{Fig:mstl_decomposition}}
	\caption{MSTL local explainer forecasts and interpretation.}
\end{figure}


Figure~\ref{Fig:mstl_forecasts} shows the same series as Figure~\ref{Fig:Tbats_forecasts}, but for the MSTL local explainer, and analogously to TBATS, we can conclude from the figure that the explainer performs as expected.
Figure~\ref{Fig:mstl_decomposition} shows the explanations for the forecasts generated by the global model using the MSTL explainer, in the form of a decomposition (again over the training set). 
Similar to the TBATS explainer, the MSTL explainer provides the daily and weekly seasonalities and the trend as the decomposed components. 

\paragraph{Example 3: Decomposition explanations using STL local explainer}

\begin{figure}[!htb]
	\centering
	\subfloat[STL local explainer forecasts]{
		\includegraphics[width=0.8\textwidth]{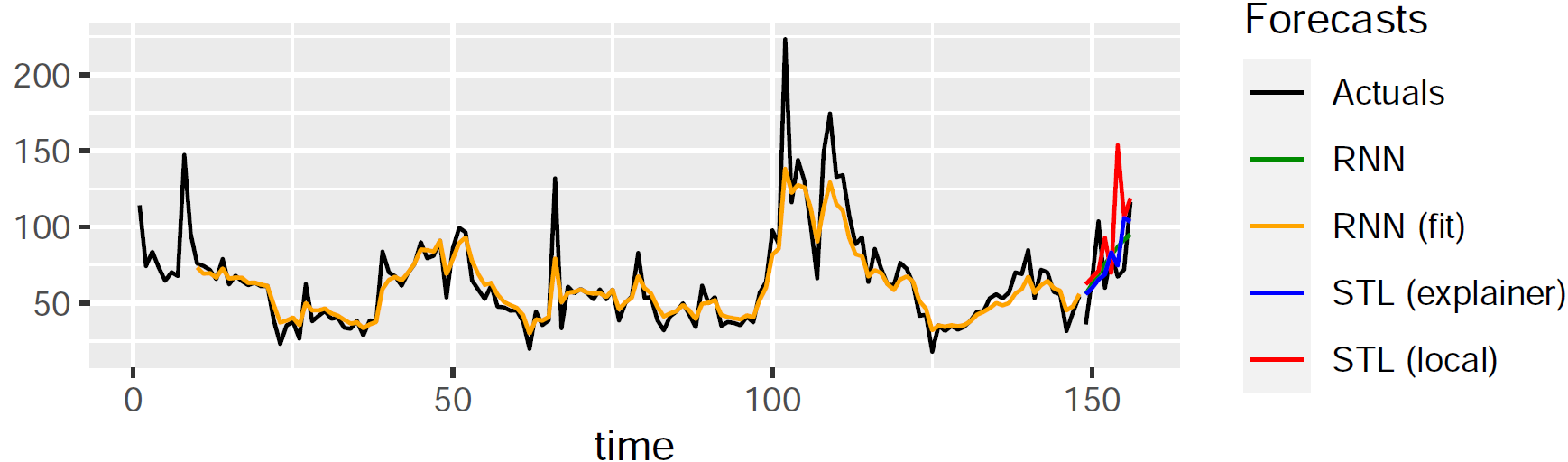}
	\label{Fig:stl_forecasts}}
		\qquad
	\subfloat[Decomposition plot for STL interpretation]{
			\includegraphics[width=0.8\textwidth]{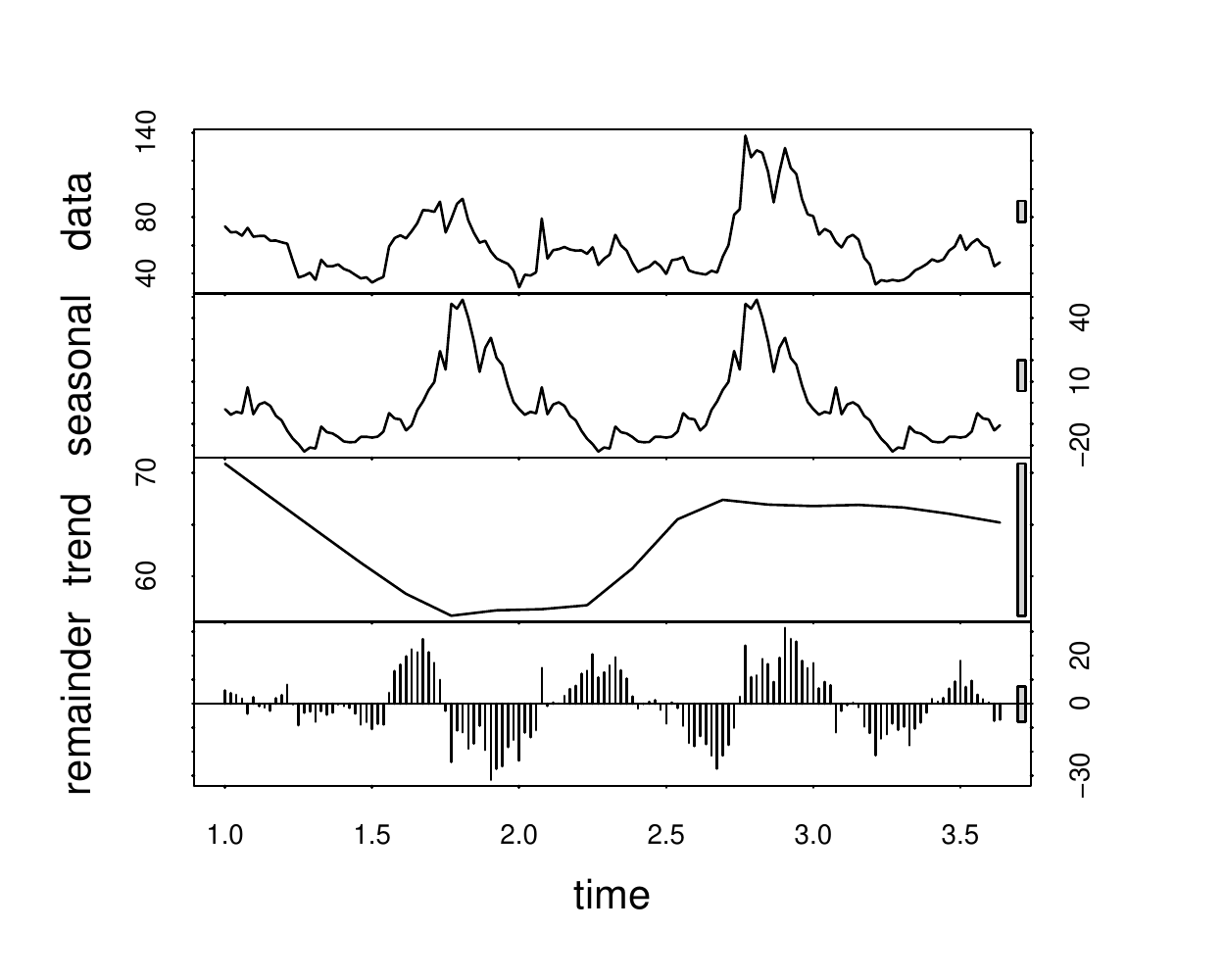}
	\label{Fig:stl_decomposition}}
	\caption{STL local explainer forecasts and interpretation.}
\end{figure}


Figure~\ref{Fig:stl_forecasts} shows the forecasts plot based on the STL model for a single time series of the Ausgrid weekly dataset.  
By observing the plot it is visible that the global RNN model outperforms the local STL model and the STL explainer model approximate the global model than the STL local model approximate the global model. 
Both of these observations conclude that the STL explainer is performing as expected.
Figure~\ref{Fig:stl_decomposition} shows the explanations for the forecasts generated by the global model using the STL explainer, as decompositions (over the training set) into the trend and the seasonality as the explanations of the global model forecasts. 


\paragraph{Example 4: Decomposition explanations using a PROPHET local explainer}

\begin{figure}[!htb]
	\centering
	\subfloat[PROPHET local explainer forecasts]{
		\includegraphics[width=0.8\textwidth]{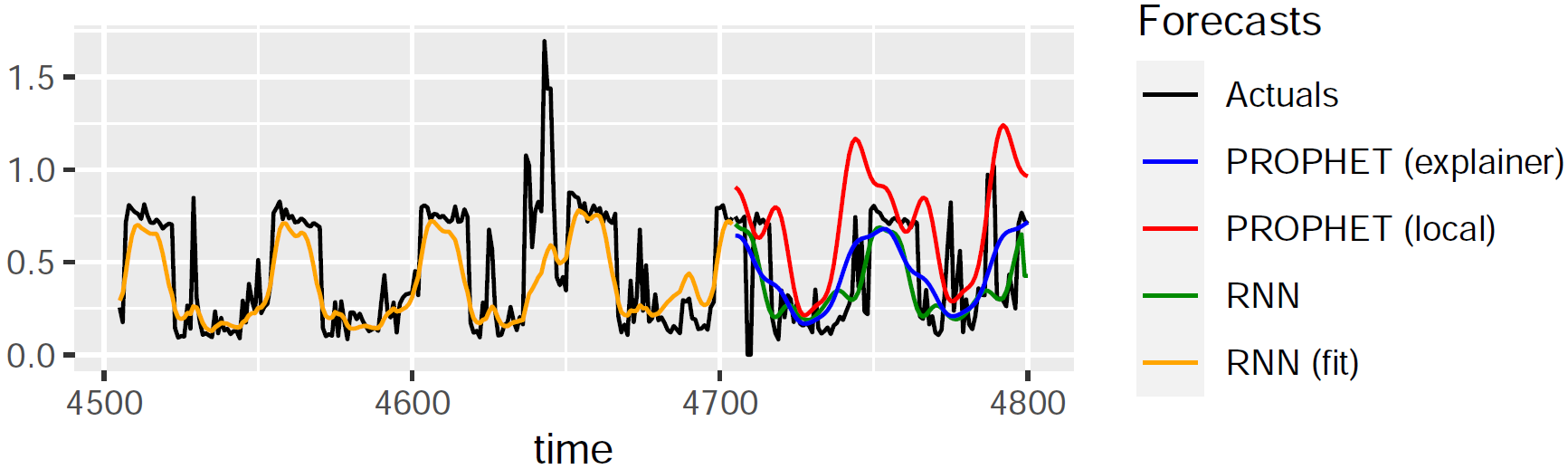}
	\label{Fig:Prophet_forecasts}}
		\qquad
	\subfloat[Decomposition plot for PROPHET interpretation]{
			\includegraphics[width=0.8\textwidth]{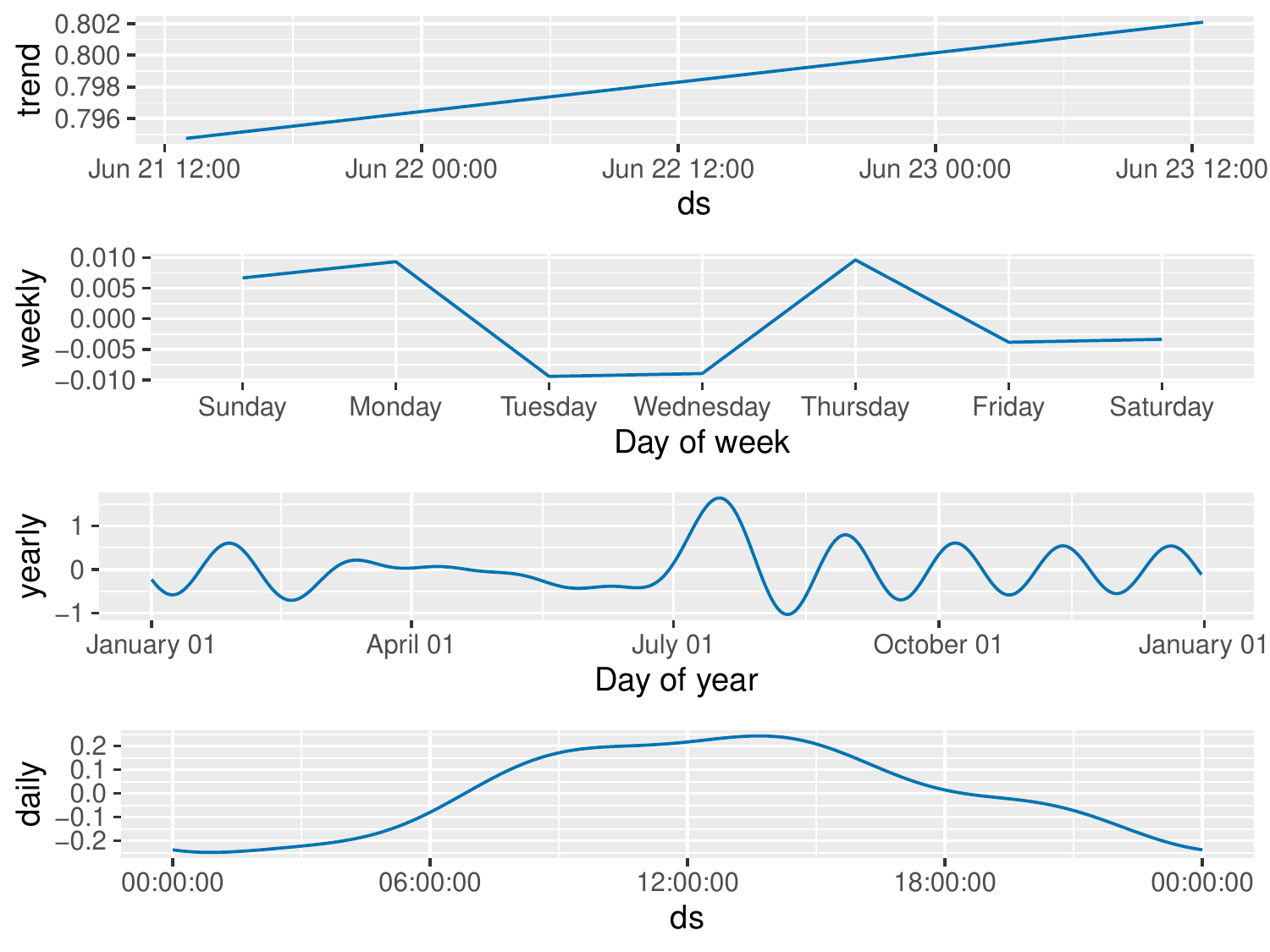}
	\label{Fig:Prophet_decomposition}}
	\caption{PROPHET local explainer forecasts and interpretation.}
\end{figure}



Figure~\ref{Fig:Prophet_forecasts} shows the forecasts plot based on the PROPHET model for a single time series of the Ausgrid half-hourly dataset.  
By observing the plot it is visible that the global RNN model outperforms the local PROPHET model and the PROPHET explainer model approximates the global model better than the PROPHET local model approximates the global model. 
Both of these observations conclude that the PROPHET explainer is performing as expected.
Figure~\ref{Fig:Prophet_decomposition} shows the explanations for the forecasts generated by the global model using the PROPHET explainer. Here, the PROPHET explainer provides daily, yearly, weekly seasonalities and the trend as components of the forecasts that can be used to attribute directly parts of the forecasts of the global model to these components.  


\subsubsection{Example explanations for bootstrapped neighbourhoods}

In contrast to the NF approach, in the NSTL and NSieve approaches, the explanations are produced by training an interpretable statistical model on the in-sample global model forecasts of the bootstrapped series.
Therefore, in this case, there are $N$ number of explainer models, $N$ being the number of bootstrapped series per original series.
In our experiments, we bootstrap 100 series from the weekly series and 50 series from the daily series. 
With $N$ explainer models instead of just one, it now also becomes important to be able to summarise the models. In the following, we illustrate this for the NSTL bootstrap and for PR and ETS as the local explainers. 


\paragraph{Example 1: Decomposition explanations using ETS local explainer}

\begin{figure}[htb]
	\centering
	\includegraphics[width=0.8\textwidth]{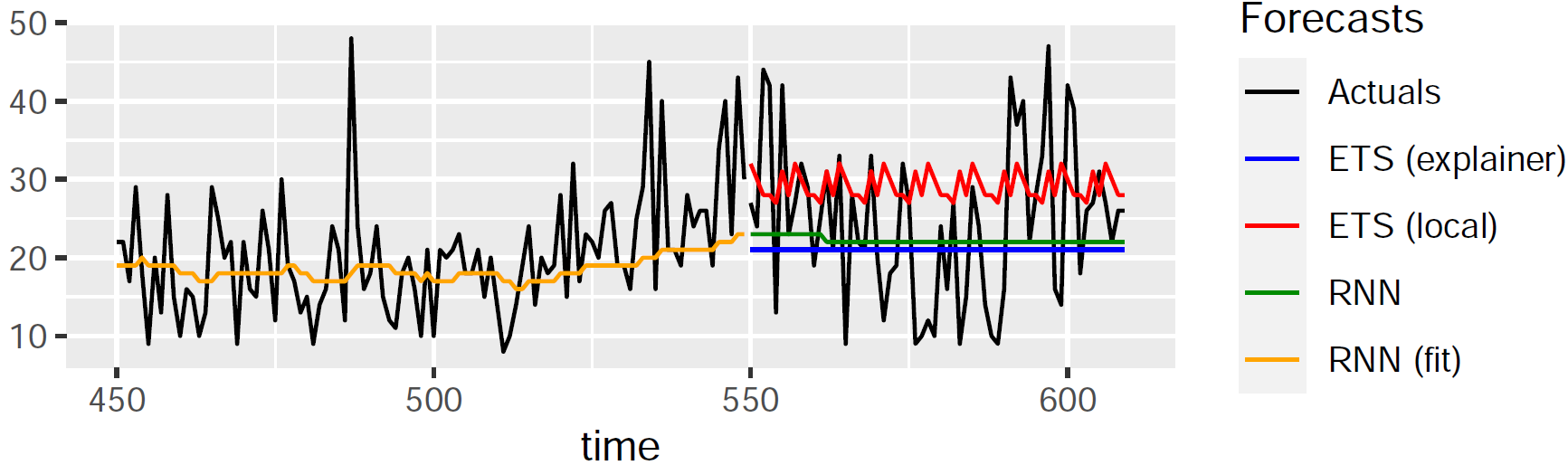}
	\caption{ETS interpretation: Forecasts plot}
	\label{Fig:ETS_forecasts_m2}
\end{figure}

Figure~\ref{Fig:ETS_forecasts_m2} shows the forecasts plot based on the ETS model for a single time series of the Kaggle web traffic daily dataset. 
Here, the forecasts of the explainer are calculated by bagging, i.e., by averaging the forecasts over the bootstrapped series. 
From the plot we can see that the global RNN model outperforms the local ETS model and the ETS explainer model approximates the global model better than the ETS local model approximates the global model. 
Both of these observations conclude that the ETS explainer is performing as expected using the NSTL explainer method. 

\begin{figure}[htb]
	\centering
	\includegraphics[width=0.8\textwidth]{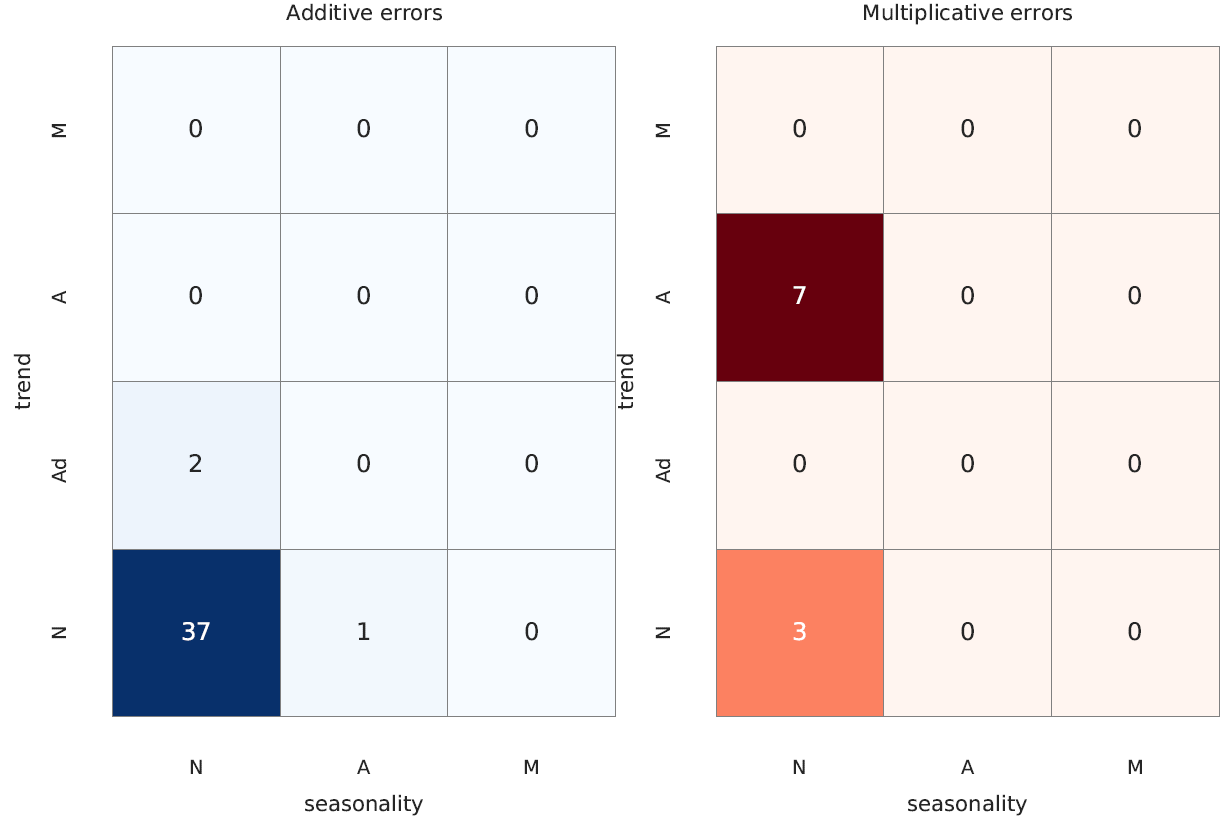}
	\caption{ETS interpretation: Chosen model summary plot}
	\label{Fig:ETS_box_m2}
\end{figure}


Figure~\ref{Fig:ETS_box_m2} summarises the resulting ETS models fitted on the bootstrapped series, following a procedure from \citet{petropoulos2018exploring}. 
As an explanation, Figure~\ref{Fig:ETS_box_m2}, shows us that the predictions are based on no seasonality and a trend that is between no trend and a linear trend, i.e., a weak linear trend can be assumed.

\paragraph{Example 2: Decomposition explanations using a PR local explainer}

\begin{figure}[htb]
	\centering
	\includegraphics[width=0.8\textwidth]{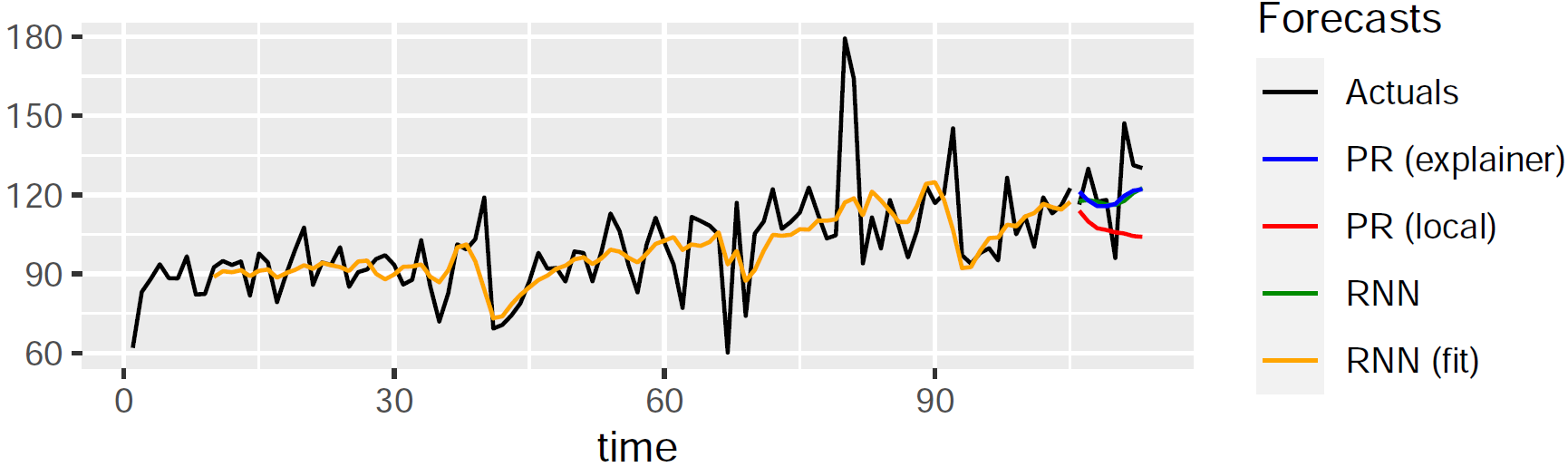}
	\caption{PR interpretation: Forecasts plot}
	\label{Fig:pooled_regression_forecasts_m2}
\end{figure}

Though PR is a global model, as outlined earlier, here we use it as a local explainer by fitting it across all bootstrapped series for a particular series to be explained.
   
Figure~\ref{Fig:pooled_regression_forecasts_m2} shows the forecasts plot based on the PR model for a single time series of the NN5 weekly dataset.  
From the plot we see that the global RNN model outperforms the PR model and the PR explainer model approximates the global model better than the (local) PR model. 

\begin{figure}[htb]
	\centering
	\includegraphics[scale=0.25]{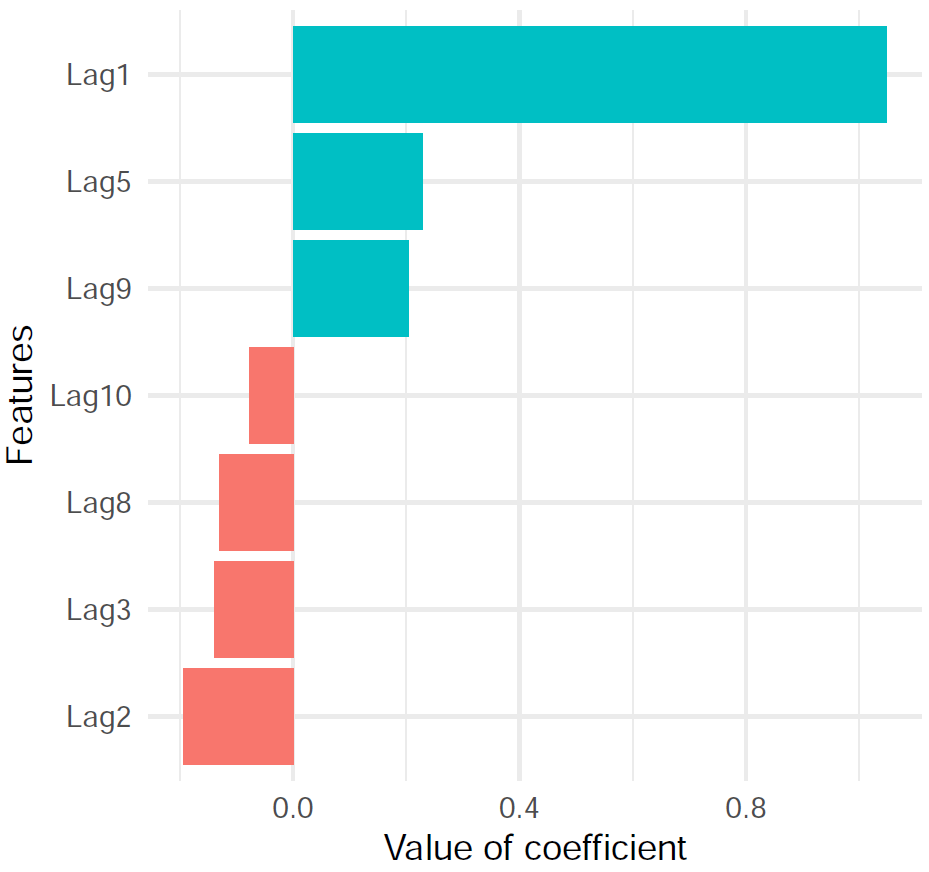}
	\caption{PR interpretation: Coefficients of significant features.}
	\label{Fig:pooled_regression_box_m2}
\end{figure}

Figure~\ref{Fig:pooled_regression_box_m2} shows the PR interpretation based on the coefficients of the regression function. 
Similar to the explanations provided by the LIME algorithm~\citep{Ribeiro2016-sp}, here we plot the coefficients in a bar graph as the explanation of feature importance for the global model forecast, as the coefficients show how much a feature contributes towrds a higher or lower prediction. 
From the figure, we can see that lag1 contributes towards higher values, whereas lag2 and lag8 contribute to lower values of the global model prediction.

\section{Conclusions and Future work}
\label{sec:conclusion}

We have presented Local Model Explanation for Forecasting (LoMEF), a model-agnostic framework for local explainability of global forecasting models. 
To the best of our knowledge, this is the first approach in the space of global model forecasting that helps to explain the global model forecasts locally in a model-agnostic way. 
It employs statistical forecasting techniques that can be deemed interpretable, in a locally defined neighbourhood to explain the global model forecast of a particular time series using interpretable components such as trend, seasonality, coefficients, and others. We have proposed three approaches to define the neighbourhood, based on bootstrapping and model fitting. 
The experiments performed both in quantitative and qualitative aspects, have shown that all three underlined methods in our LoMEF framework lead to explainers that approximate the global model forecast well, and that lead to comprehensible explanations.




\bibliography{refs}

\appendix
\section{Fidelity and accuracy measures for NF method}

\begin{table}[H]
	\centering
	\caption{Mean and median performance measures based on the MASE error measure for datasets on different local explainer models. Asterisks indicate statistically significant results, boldface font indicates negative values (which are desirable).} 
	\label{tab:mm_mase}
	\scalebox{0.7}{
		\begin{tabular}{llcccccc}
			\hline
			Dataset 
			&\multicolumn{1}{p{1.5cm}}{\centering  Local\\ Explainer} 
			& Fidelity\_Actual 
			& Fidelity\_Local 
			& \multicolumn{1}{p{2.0cm}}{\centering Fidelity\_with\\\_Explainer}
			&\multicolumn{1}{p{2.0cm}}{\centering Acc\_Global\_\\LocalModel} 
			&\multicolumn{1}{p{2.2cm}}{\centering Acc\_Explainer and \\ Local Model} 
			& \multicolumn{1}{p{2.2cm}}{\centering Acc\_Explainer and \\Global Model}\\ 
			\hline
			\multicolumn{8}{c}{Mean Values}\\
			 \hline
			NN5W& PROPHET & \textbf{-0.375}\anote & \textbf{-0.146}\anote & \textbf{-0.086} & \textbf{-0.033} & \textbf{-0.005} & 0.028 \\ 
			& DHR\_ARIMA & \textbf{-0.439}\anote & \textbf{-0.147}\anote & \textbf{-0.086} & \textbf{-0.026} & \textbf{-0.017} & 0.009 \\ 
			& TBATS &\textbf{ -0.503}\anote & \textbf{-0.246}\anote & \textbf{-0.130} & \textbf{-0.187}\anote & \textbf{-0.140}\anote & 0.047 \\ 
			AusGridW & PROPHET & \textbf{-0.268}\anote & \textbf{-0.083 }\anote& 0.272 & \textbf{-0.126}\anote & \textbf{-0.063}\anote & 0.063 \\ 
			& DHR\_ARIMA & \textbf{-0.335}\anote & \textbf{-0.128}\anote & 0.179 & \textbf{-0.099} &\textbf{ -0.066}\anote & 0.033 \\ 
			& TBATS & \textbf{-0.360}\anote &\textbf{ -0.087}\anote & 0.075 & \textbf{-0.124}\anote & \textbf{-0.046} & 0.078 \\ 
			& STL\_ETS & \textbf{-0.302}\anote & -\textbf{0.183}\anote & 0.049 & \textbf{-0.186}\anote & \textbf{-0.127}\anote & 0.058 \\ 
			WebTrafficD & ETS & \textbf{-1.101}\anote & \textbf{-0.773}\anote & \textbf{-0.739}\anote & \textbf{-0.431} & \textbf{-0.430} & 0.001 \\ 
			& THETA & \textbf{-1.110}\anote & \textbf{-0.424}\anote & \textbf{-0.394}\anote & \textbf{-0.157}\anote & \textbf{-0.155}\anote & 0.003 \\ 
			SFTrafficH & TBATS & \textbf{-0.364}\anote & \textbf{-0.480 }\anote& \textbf{-0.413}\anote & \textbf{-0.170}\anote & \textbf{-0.182}\anote & \textbf{-0.013 }\\ 
			& PROPHET & \textbf{-0.232}\anote & \textbf{-0.354 }\anote& \textbf{-0.194}\anote & \textbf{-0.453}\anote & \textbf{-0.298}\anote & 0.155 \\ 
			& DHR\_ARIMA & \textbf{-0.262}\anote & \textbf{-0.547}\anote & \textbf{-0.274}\anote & \textbf{-0.291}\anote & \textbf{-0.308}\anote & \textbf{-0.018} \\ 
			& MSTL\_ETS & \textbf{-0.032} & \textbf{-0.366}\anote &\textbf{ -0.210 }\anote&\textbf{ -0.334}\anote & \textbf{-0.183}\anote & 0.152 \\ 
			AusGridHH & TBATS & \textbf{-0.462}\anote & \textbf{-0.510}\anote & \textbf{-0.446}\anote & \textbf{-0.285 }\anote& \textbf{-0.291} \anote& \textbf{-0.006} \\ 
			& PROPHET & \textbf{-0.337}\anote & \textbf{-0.680}\anote & \textbf{-0.578}\anote & \textbf{-0.533}\anote & \textbf{-0.449}\anote & 0.083 \\ 
			& DHR\_ARIMA &\textbf{ -0.408}\anote & \textbf{-0.347}\anote & \textbf{-0.160 }& \textbf{-0.155 }\anote& \textbf{-0.142}\anote & 0.013 \\ 
			& MSTL\_ETS & 0.110 & \textbf{-0.393}\anote & \textbf{-0.595}\anote & \textbf{-0.645}\anote & \textbf{-0.197} & 0.448 \\ 
	
			\hline
			\multicolumn{8}{c}{Median Values}\\
			  \hline
			NN5W& PROPHET & \textbf{-0.375}\anote & \textbf{-0.116 }\anote& \textbf{-0.057} & \textbf{-0.049} & \textbf{-0.030} & \textbf{-0.005} \\ 
			& DHR\_ARIMA & \textbf{-0.435}\anote & \textbf{-0.108}\anote &\textbf{ -0.071 }& \textbf{-0.029} & \textbf{-0.021} & 0.012 \\ 
			& TBATS & \textbf{-0.445}\anote & \textbf{-0.219}\anote & \textbf{-0.067} & \textbf{-0.114}\anote & \textbf{-0.094}\anote & 0.046 \\ 
			AusGridW& PROPHET & \textbf{-0.174 }\anote& \textbf{-0.081}\anote & 0.172 & \textbf{-0.103}\anote & \textbf{-0.063}\anote & 0.041 \\ 
			& DHR\_ARIMA & \textbf{-0.210}\anote & \textbf{-0.109}\anote & 0.070 & \textbf{-0.067} & \textbf{-0.052}\anote & 0.044 \\ 
			& TBATS & \textbf{-0.215}\anote & \textbf{-0.076}\anote & 0.057 & \textbf{-0.064}\anote & \textbf{-0.030} & 0.050 \\ 
			& STL\_ETS & \textbf{-0.196}\anote & \textbf{-0.170}\anote & \textbf{-0.042} & \textbf{-0.152}\anote & \textbf{-0.124}\anote & 0.055 \\ 
			WebTrafficD & ETS & \textbf{-0.713 }\anote& \textbf{-0.253}\anote & \textbf{-0.245}\anote & \textbf{-0.057} & \textbf{-0.050} & 0.000 \\ 
			& THETA & \textbf{-0.714}\anote & \textbf{-0.204}\anote & \textbf{-0.189}\anote & \textbf{-0.032}\anote & \textbf{-0.029}\anote & 0.000 \\ 
			SFTrafficH & TBATS & \textbf{-0.298}\anote & \textbf{-0.348 }\anote& \textbf{-0.289}\anote & \textbf{-0.071}\anote & \textbf{-0.092 }\anote& \textbf{-0.030} \\ 
			& PROPHET & \textbf{-0.183}\anote & \textbf{-0.331}\anote & \textbf{-0.190}\anote & \textbf{-0.420}\anote & \textbf{-0.282 }\anote& 0.120 \\ 
			& DHR\_ARIMA & \textbf{-0.218}\anote & \textbf{-0.470 }\anote& \textbf{-0.206}\anote & \textbf{-0.208}\anote & \textbf{-0.242}\anote &\textbf{ -0.017} \\ 
			& MSTL\_ETS & \textbf{-0.052} & \textbf{-0.362}\anote & \textbf{-0.200 }\anote & \textbf{-0.285}\anote & \textbf{-0.187}\anote & 0.071 \\ 
			AusGridHH & TBATS & \textbf{-0.339}\anote & \textbf{-0.323}\anote & \textbf{-0.252}\anote & \textbf{-0.144}\anote & \textbf{-0.167}\anote & \textbf{-0.006} \\ 
			& PROPHET & \textbf{-0.211}\anote & \textbf{-0.553}\anote & \textbf{-0.435}\anote & \textbf{-0.433}\anote & \textbf{-0.351}\anote & 0.048 \\ 
			& DHR\_ARIMA & \textbf{-0.275}\anote & \textbf{-0.227}\anote & \textbf{-0.055} & \textbf{-0.129}\anote & \textbf{-0.106}\anote & 0.012 \\ 
			& MSTL\_ETS& \textbf{-0.166} & \textbf{-0.514}\anote & \textbf{-0.420}\anote & \textbf{-0.485}\anote & \textbf{-0.354} & 0.085 \\ 
			\hline
		\end{tabular}
	}
\end{table}

\begin{table}[H]
	\vspace{-2cm}
	\centering
	\caption{Mean and median performance measures based on the MAE error measure for datasets on different local explainer models. Asterisks indicate statistically significant results, boldface font indicates negative values (which are desirable).} 
	\label{tab:mm_mae}
	\scalebox{0.7}{
		\begin{tabular}{llcccccc}
			\hline
			Dataset 
			&\multicolumn{1}{p{1.5cm}}{\centering  Local\\ Explainer} 
			& Fidelity\_Actual 
			& Fidelity\_Local 
			& \multicolumn{1}{p{2.0cm}}{\centering Fidelity\_with\\\_Explainer}
			&\multicolumn{1}{p{2.0cm}}{\centering Acc\_Global\_\\LocalModel} 
			&\multicolumn{1}{p{2.2cm}}{\centering Acc\_Explainer and \\ Local Model} 
			& \multicolumn{1}{p{2.2cm}}{\centering Acc\_Explainer and \\Global Model}\\ 
			\hline
			\multicolumn{8}{c}{Mean Values}\\
			\hline
			NN5W& PROPHET &\textbf{ -6.454}\anote & \textbf{-2.794}\anote &\textbf{ -1.427} & \textbf{-0.683} & \textbf{-0.131} & 0.552 \\ 
			& DHR\_ARIMA & \textbf{-7.673}\anote & \textbf{-2.793}\anote & \textbf{-1.651} & \textbf{-0.586} & \textbf{-0.406} & 0.180 \\ 
			& TBATS & \textbf{-8.419}\anote & \textbf{-4.555}\anote & \textbf{-2.179} & \textbf{-3.332}\anote & \textbf{-2.392}\anote & 0.940 \\ 
			AusGridW& PROPHET & \textbf{-6.093}\anote & \textbf{-2.643}\anote & 7.969 &\textbf{ -4.649}\anote &\textbf{ -2.045}\anote & 2.604 \\ 
			& DHR\_ARIMA & \textbf{-8.242}\anote & \textbf{-3.282}\anote & 5.105 & \textbf{-3.142} & \textbf{-1.689} & 1.453 \\ 
			& TBATS & \textbf{-9.319}\anote & \textbf{-1.956} & 1.915 & \textbf{-3.000}\anote & \textbf{-0.568} & 2.432 \\ 
			& STL\_ETS & \textbf{-7.244}\anote & \textbf{-4.689}\anote & 2.067 & \textbf{-5.748}\anote & \textbf{-3.518}\anote & 2.229 \\ 
			WebTrafficD & ETS &\textbf{ -10.887}\anote &\textbf{ -8.669}\anote & \textbf{-7.772}\anote &\textbf{ -5.156}\anote & \textbf{-5.215}\anote &\textbf{ -0.059} \\ 
			& THETA & \textbf{-11.045 }\anote& \textbf{-5.904}\anote &\textbf{ -5.250}\anote & \textbf{-2.798} & \textbf{-2.796} & 0.001 \\ 
			SFTrafficH & TBATS & \textbf{-0.005}\anote & \textbf{-0.006}\anote &\textbf{ -0.005}\anote & \textbf{-0.002}\anote & \textbf{-0.002}\anote & 0.000 \\ 
			& PROPHET & \textbf{-0.003}\anote & \textbf{-0.005}\anote & \textbf{-0.003}\anote &\textbf{ -0.006}\anote &\textbf{ -0.004}\anote & 0.002 \\ 
			& DHR\_ARIMA & \textbf{-0.004}\anote & \textbf{-0.008}\anote & \textbf{-0.004}\anote& \textbf{-0.004}\anote & \textbf{-0.004}\anote & 0.000 \\ 
			& MSTL\_ETS & 0.000 &\textbf{ -0.005}\anote & \textbf{-0.002}\anote &\textbf{ -0.005}\anote &\textbf{ -0.002}\anote & 0.003 \\ 
			AusGridHH & TBATS &\textbf{ -0.075}\anote &\textbf{ -0.115}\anote &\textbf{ -0.102}\anote & \textbf{-0.078}\anote &\textbf{ -0.078}\anote & 0.000 \\ 
			& PROPHET & -0.046\anote & \textbf{-0.142}\anote & \textbf{-0.121}\anote & \textbf{-0.122}\anote & \textbf{-0.103}\anote & 0.019 \\ 
			& DHR\_ARIMA &\textbf{ -0.061}\anote &\textbf{ -0.056}\anote & \textbf{-0.016}\anote & \textbf{-0.035}\anote & \textbf{-0.029}\anote & 0.005 \\ 
			& MSTL\_ETS & 0.055 & \textbf{-0.087} &\textbf{ -0.132}\anote &\textbf{ -0.165}\anote & \textbf{-0.061} & 0.104 \\ 
			\hline
			\multicolumn{8}{c}{Median Values}\\
			\hline
			NN5W& PROPHET & \textbf{-6.537}\anote &\textbf{ -1.951}\anote & \textbf{-0.824} & \textbf{-0.933} & \textbf{-0.405} & \textbf{-0.059} \\ 
			& DHR\_ARIMA & \textbf{-6.387}\anote & \textbf{-2.134}\anote & \textbf{-1.281} & \textbf{-0.602} & \textbf{-0.457} & 0.213 \\ 
			& TBATS & \textbf{-6.851}\anote &\textbf{} \textbf{-3.670}\anote & \textbf{-1.090} & \textbf{-2.241}\anote & \textbf{-1.319}\anote & 0.619 \\ 
			AusGridW& PROPHET & \textbf{-4.254}\anote & \textbf{-1.820}\anote & 3.998 & \textbf{-2.190}\anote & \textbf{-1.384}\anote & 1.043 \\ 
			& DHR\_ARIMA & \textbf{-4.835}\anote & \textbf{-2.641}\anote & 2.031 & \textbf{-1.294} & \textbf{-1.163} & 0.947 \\ 
			& TBATS & \textbf{-5.571}\anote & \textbf{-1.952} & 1.347 & \textbf{-1.655}\anote & \textbf{-0.674} & 1.390 \\ 
			& STL\_ETS & \textbf{-5.354}\anote & \textbf{-3.839}\anote & \textbf{-0.828} &\textbf{ -3.570}\anote & \textbf{-2.899}\anote & 1.507 \\ 
			WebTrafficD & ETS & \textbf{-5.850}\anote & \textbf{-2.000}\anote &\textbf{ -2.000}\anote & \textbf{-0.467}\anote & \textbf{-0.433}\anote & 0.000 \\ 
			& THETA & \textbf{-5.933}\anote &\textbf{ -1.617}\anote & \textbf{-1.533}\anote &\textbf{ -0.283} & \textbf{-0.250} & 0.000 \\ 
			SFTrafficH & TBATS &\textbf{ -0.004}\anote & \textbf{-0.005}\anote & \textbf{-0.004}\anote & \textbf{-0.001}\anote & \textbf{-0.001}\anote & 0.000 \\ 
			& PROPHET &\textbf{ -0.003}\anote &\textbf{ -0.004}\anote & \textbf{-0.003}\anote & \textbf{-0.006}\anote &\textbf{ -0.004}\anote & 0.002 \\ 
			& DHR\_ARIMA & \textbf{-0.003}\anote & \textbf{-0.006}\anote & \textbf{-0.003}\anote & \textbf{-0.003}\anote &\textbf{ -0.003}\anote & 0.000 \\ 
			& MSTL\_ETS & \textbf{-0.001} & \textbf{-0.005}\anote & \textbf{-0.003}\anote & \textbf{-0.004}\anote & \textbf{-0.003}\anote & 0.001 \\ 
			AusGridHH & TBATS & \textbf{-0.064}\anote & \textbf{-0.059}\anote & \textbf{-0.047}\anote & \textbf{-0.028}\anote & \textbf{-0.031}\anote & \textbf{-0.001} \\ 
			& PROPHET &\textbf{-0.041}\anote & \textbf{-0.092}\anote & \textbf{-0.079}\anote & \textbf{-0.080}\anote & \textbf{-0.068}\anote & 0.009 \\ 
			& DHR\_ARIMA & \textbf{-0.054}\anote & \textbf{-0.046}\anote & \textbf{-0.012}\anote & \textbf{-0.026}\anote & \textbf{-0.021}\anote & 0.002 \\ 
			& MSTL\_ETS & \textbf{-0.029} & \textbf{-0.087} &\textbf{ -0.075}\anote & \textbf{-0.094}\anote & \textbf{-0.058} & 0.014 \\ 
			\hline
		\end{tabular}
	}
\end{table}
	
\section{Fidelity and accuracy measures for NSTL method}

\begin{table}[H]
	\centering
	\caption{Mean and median performance measures based on MASE error measures of the datasets on different local explainer models using NSTL.} 
	\label{tab:mm_mase_bm1}
	\scalebox{0.75}{
		\begin{tabular}{llcccccc}
			\hline
			Dataset 
			&\multicolumn{1}{p{1.5cm}}{\centering  Local\\ Explainer} 
			& Fidelity\_Actual 
			& Fidelity\_Local 
			& \multicolumn{1}{p{2.0cm}}{\centering Fidelity\_with\\\_Explainer}
			&\multicolumn{1}{p{2.0cm}}{\centering Acc\_Global\_\\LocalModel} 
			&\multicolumn{1}{p{2.2cm}}{\centering Acc\_Explainer and \\ Local Model} 
			& \multicolumn{1}{p{2.2cm}}{\centering Acc\_Explainer and \\Global Model}\\ 
			\hline
			\multicolumn{8}{c}{Mean Values}\\
			\hline
			NN5W & PR & \textbf{-0.236}\anote & 0.145 & 0.060 & \textbf{-0.087}\anote & \textbf{-0.052 }& 0.034 \\ 
				& ETS & - & - & -  & - &-&-\\
			WebTraffic100D & PR& \textbf{-0.862}\anote & \textbf{-0.120} & \textbf{-0.167 }\anote & \textbf{-0.081} & \textbf{-0.091} & \textbf{-0.010} \\
			&ETS & \textbf{-0.929}\anote & \textbf{-0.735 }\anote & \textbf{-0.728}\anote & \textbf{-0.479} & \textbf{-0.421 }& 0.058 \\ 

			\hline
			\multicolumn{8}{c}{Median Values}\\ 
			 \hline
			NN5W & PR &\textbf{ -0.234}\anote & 0.013 & 0.051 & \textbf{-0.058}\anote & \textbf{-0.057} & 0.018 \\ 
			& ETS & - & - & -  & - &-&-\\
			WebTraffic100D 	& PR& \textbf{-0.603}\anote & \textbf{-0.114} & \textbf{-0.116}\anote & \textbf{-0.032} & \textbf{-0.023} & 0.000 \\ 
			& ETS & \textbf{-0.631}\anote & \textbf{-0.290}\anote & \textbf{-0.303}\anote & \textbf{-0.145} & \textbf{-0.052} & 0.051 \\ 
		
			\hline
		\end{tabular}
	}
\end{table}

\begin{table}[H]
	\centering
	\caption{Mean and median performance measures based on MAE error measures of the datasets on different local explainer models using NSTL.} 
	\label{tab:mm_mae_bm1}
	\scalebox{0.75}{
		\begin{tabular}{llcccccc}
			\hline
			Dataset 
			&\multicolumn{1}{p{1.5cm}}{\centering  Local\\ Explainer} 
			& Fidelity\_Actual 
			& Fidelity\_Local 
			& \multicolumn{1}{p{2.0cm}}{\centering Fidelity\_with\\\_Explainer}
			&\multicolumn{1}{p{2.0cm}}{\centering Acc\_Global\_\\LocalModel} 
			&\multicolumn{1}{p{2.2cm}}{\centering Acc\_Explainer and \\ Local Model} 
			& \multicolumn{1}{p{2.2cm}}{\centering Acc\_Explainer and \\Global Model}\\ 
			\hline
			\multicolumn{8}{c}{Mean Values}\\
			\hline
			NN5W & PR &\textbf{ -3.879}\anote & 1.072 & \textbf{-0.277} & \textbf{-2.556} & \textbf{-2.033} & 0.523 \\ 
				& ETS & - & - & -  & - &-&-\\
			WebTraffic100D	& PR & \textbf{-7.896}\anote & \textbf{-1.489}\anote &\textbf{ -1.550}\anote & \textbf{-0.924}\anote &\textbf{ -0.807} & 0.118 \\ 
			 & ETS & \textbf{-8.826}\anote & \textbf{-9.175}\anote &\textbf{ -9.126}\anote & \textbf{-5.470} & \textbf{-5.511} & \textbf{-0.041} \\ 
		
			\hline
			\multicolumn{8}{c}{Median Values}\\ 
			\hline
			NN5W & PR & \textbf{-4.287}\anote & 0.279 & 0.886 &\textbf{ -0.787} & \textbf{-0.773} & 0.283 \\ 
				& ETS & - & - & -  & - &-&-\\
			WebTraffic100D& PR & \textbf{-5.458}\anote &\textbf{ -0.732}\anote & \textbf{-0.915}\anote & \textbf{-0.196}\anote & \textbf{-0.151} & 0.000 \\ 
			 & ETS &\textbf{ -6.167}\anote & \textbf{-2.150}\anote & \textbf{-2.917}\anote &\textbf{ -0.633} & \textbf{-0.442} & 0.000 \\ 
			
			\hline
		\end{tabular}
	}
\end{table}

\section{Fidelity and accuracy measures for NSieve method}

\begin{table}[H]
	\centering
	\caption{Mean and median performance measures based on MASE error measures of the datasets on different local explainer models using NSieve approach.} 
	\label{tab:mm_mase_bm2}
	\scalebox{0.75}{
		\begin{tabular}{llcccccc}
			\hline
			Dataset 
			&\multicolumn{1}{p{1.5cm}}{\centering  Local\\ Explainer} 
			& Fidelity\_Actual 
			& Fidelity\_Local 
			& \multicolumn{1}{p{2.0cm}}{\centering Fidelity\_with\\\_Explainer}
			&\multicolumn{1}{p{2.0cm}}{\centering Acc\_Global\_\\LocalModel} 
			&\multicolumn{1}{p{2.2cm}}{\centering Acc\_Explainer and \\ Local Model} 
			& \multicolumn{1}{p{2.2cm}}{\centering Acc\_Explainer and \\Global Model}\\ 
			\hline
			\multicolumn{8}{c}{Mean Values}\\
			\hline
			NN5W & PR & \textbf{-0.282}\anote & 0.099 & 0.176 & \textbf{-0.087}\anote & 0.033 & 0.120 \\ 
				& ETS & - & - & -  & - &-&-\\
			WebTraffic100D & PR& \textbf{-0.792 }\anote& \textbf{-0.013} & \textbf{-0.108} & \textbf{-0.086 }& \textbf{-0.004} & 0.082 \\ 
			& ETS & \textbf{-0.781}\anote & \textbf{-0.528} &\textbf{ -0.753}\anote & \textbf{-0.420} & \textbf{-0.325} & 0.094 \\ 
			
			\hline
			\multicolumn{8}{c}{Median Values}\\  
			 \hline
			NN5W & PR & \textbf{-0.248 }\anote& 0.059 & 0.164 & \textbf{-0.058 }\anote& 0.028 & 0.097 \\
				& ETS & - & - & -  & - &-&-\\ 
			WebTraffic100D& PR & \textbf{-0.591}\anote & \textbf{-0.045} & \textbf{-0.111} & \textbf{-0.027} & \textbf{-0.012} & 0.005 \\ 
			 & ETS & \textbf{-0.557}\anote & \textbf{-0.134} & \textbf{-0.292 }\anote & \textbf{-0.069} & \textbf{-0.031} & 0.009 \\ 
			& PR & \textbf{-0.591}\anote & \textbf{-0.045} & \textbf{-0.111} & \textbf{-0.027} & \textbf{-0.012} & 0.005 \\ 
			\hline
		\end{tabular}
	}
\end{table}

\begin{table}[H]
	\centering
	\caption{Mean and median performance measures based on MAE error measures of the datasets on different local explainer models using NSieve approach.} 
	\label{tab:mm_mae_bm2}
	\scalebox{0.75}{
		\begin{tabular}{llcccccc}
			\hline
			Dataset 
			&\multicolumn{1}{p{1.5cm}}{\centering  Local\\ Explainer} 
			& Fidelity\_Actual 
			& Fidelity\_Local 
			& \multicolumn{1}{p{2.0cm}}{\centering Fidelity\_with\\\_Explainer}
			&\multicolumn{1}{p{2.0cm}}{\centering Acc\_Global\_\\LocalModel} 
			&\multicolumn{1}{p{2.2cm}}{\centering Acc\_Explainer and \\ Local Model} 
			& \multicolumn{1}{p{2.2cm}}{\centering Acc\_Explainer and \\Global Model}\\ 
			\hline
			\multicolumn{8}{c}{Mean Values}\\
			\hline
			NN5W & PR & \textbf{-4.171}\anote & 0.781 & 3.188 & \textbf{-2.556} & \textbf{-0.068} & 2.489 \\ 
				& ETS & - & - & -  & - &-&-\\
			WebTraffic100D	& PR & \textbf{-6.401}\anote & 0.013 &\textbf{ -1.280} & \textbf{-0.923}\anote &\textbf{ -0.321} & 0.602 \\ 
			 & ETS &\textbf{ -6.428} \anote& \textbf{-6.777} & \textbf{-9.276}\anote &\textbf{ -5.470} & \textbf{-4.842} & 0.627 \\ 
		
			\hline
			\multicolumn{8}{c}{Median Values}\\  
			\hline
			NN5W & PR & \textbf{-4.271}\anote & 0.976 & 2.567\textbf{} & \textbf{-0.787} & 0.536 & 1.390 \\ 
				& ETS & - & - & -  & - &-&-\\
			WebTraffic100D	& PR & \textbf{-4.850}\anote & \textbf{-0.258} & \textbf{-0.817} & \textbf{-0.183}\anote & \textbf{-0.067} & 0.058 \\ 
			 & ETS & \textbf{-4.833}\anote & \textbf{-1.008} & \textbf{-2.567}\anote & \textbf{-0.633} & \textbf{-0.300} & 0.063 \\ 
		
			\hline
		\end{tabular}
	}
\end{table}

\section{Statistical Significance of the Results of NF explainer}

\begin{table}[H]
	\centering
	\caption{Unadjusted $p$-values for Mean RMSE measures of the datasets for different local explainer models. Significant values that are significant according to the Bonferroni-corrected significance level of $1.6\times 10^{-4}$ are shown in boldface.} 
	\label{tab:mean_rmse_stat}
	\scalebox{0.7}{
		\begin{tabular}{llcccccc}
			\hline
			Dataset 
			&\multicolumn{1}{p{1.5cm}}{\centering  Local\\ Explainer} 
			& Fidelity\_Actual 
			& Fidelity\_Local 
			& \multicolumn{1}{p{2.0cm}}{\centering Fidelity\_with\\\_Explainer}
			&\multicolumn{1}{p{2.0cm}}{\centering Acc\_Global\_\\LocalModel} 
			&\multicolumn{1}{p{2.2cm}}{\centering Acc\_Explainer and \\ Local Model} 
			& \multicolumn{1}{p{2.2cm}}{\centering Acc\_Explainer and \\Global Model}\\ 
			\hline
			NN5W& PROPHET & \textbf{2.190E-18} & \textbf{2.435E-08} & 2.134E-04 & 1.436E-01 & 3.476E-01 & 8.794E-01 \\ 
			& DHR\_ARIMA & \textbf{2.226E-23 }& \textbf{5.478E-09} & \textbf{2.519E-05} & 2.538E-01 & 1.976E-01 & 4.720E-01 \\ 
			& TBATS & \textbf{5.584E-23} & \textbf{6.950E-10} & 1.514E-03 & \textbf{6.954E-07} & \textbf{4.983E-05} & 9.962E-01 \\ 
			AusGridW & ETS & \textbf{1.529E-06} & \textbf{7.044E-10 }& 1.000E+00 & \textbf{4.286E-13 }& 2.204E-04 & 1.000E+00 \\ 
			& PROPHET & \textbf{5.439E-12} & \textbf{5.639E-09} & 1.000E+00 & 1.991E-04 & \textbf{3.034E-06} & 9.791E-01 \\ 
			&DHR\_ARIMA & \textbf{4.533E-17 }& \textbf{5.756E-11} & 1.000E+00 & 1.506E-03 & 1.733E-04 & 9.324E-01 \\ 
			& TBATS & \textbf{3.698E-19 }& 5.226E-04 & 9.916E-01 & 1.435E-04 & 1.571E-01 & 9.985E-01 \\ 
			& STL\-ETS &\textbf{ 7.380E-14} & \textbf{1.283E-27} & 8.121E-01 & \textbf{2.548E-09} & \textbf{2.475E-15} & 9.908E-01 \\ 
			WebTrafficD & ETS & 4.244E-16 & \textbf{1.480E-10} & \textbf{1.212E-10} & 3.546E-04 & 3.240E-04 & 4.947E-01 \\ 
			& THETA & \textbf{3.750E-16} & \textbf{4.276E-06} & \textbf{3.301E-06} & 2.313E-02 & 2.419E-02 & 6.443E-01 \\ 
			SFTrafficH & TBATS & \textbf{1.068E-78} & \textbf{1.729E-121} &\textbf{ 1.070E-84 }& \textbf{5.388E-17} &\textbf{ 1.623E-28} &\textbf{ 1.447E-12} \\ 
			& PROPHET & \textbf{1.806E-51} & \textbf{7.091E-182 }& \textbf{1.461E-68} & \textbf{2.253E-189} & \textbf{8.686E-150} & 1.000E+00 \\ 
			& DHR\_ARIMA  &\textbf{ 1.984E-51 }&\textbf{ 3.659E-166} & \textbf{6.729E-54 }&\textbf{ 2.720E-60 }& \textbf{4.436E-99 }& \textbf{1.176E-15 }\\ 
			& MSTL\-ETS & 2.585E-04 & \textbf{3.117E-77} & \textbf{2.843E-18} & \textbf{2.189E-66} &\textbf{ 3.198E-37} & 1.000E+00 \\ 
			AusGridHH & TBATS & \textbf{1.245E-67 }& \textbf{6.560E-19} & \textbf{1.973E-15} & \textbf{1.983E-10} & \textbf{9.561E-11} & 1.565E-01 \\ 
			& PROPHET & \textbf{1.642E-49} & \textbf{5.192E-47} &\textbf{ 7.933E-37} & \textbf{2.166E-31} & \textbf{1.721E-30 }& 1.000E+00 \\ 
			& DHR\_ARIMA & \textbf{5.191E-60} & \textbf{4.320E-57 }& \textbf{3.081E-11} & \textbf{1.910E-11} & \textbf{5.424E-18} & 4.452E-01 \\ 
			& MSTL\-ETS & 7.082E-01 & \textbf{1.641E-06} & \textbf{2.203E-20} & \textbf{1.884E-27 }& 4.016E-04 & 1.000E+00 \\ 
			\hline
		\end{tabular}
	}
\end{table}

\begin{table}[H]
	\centering
	\caption{Unadjusted $p$-values for Mean MASE measures of the datasets for different local explainer models. Significant values that are significant according to the Bonferroni-corrected significance level of $1.6\times 10^{-4}$ are shown in boldface.} 
	\label{tab:mean_mase_stat}
	\scalebox{0.7}{
		\begin{tabular}{llcccccc}
			\hline
			Dataset 
			&\multicolumn{1}{p{1.5cm}}{\centering  Local\\ Explainer} 
			& Fidelity\_Actual 
			& Fidelity\_Local 
			& \multicolumn{1}{p{2.0cm}}{\centering Fidelity\_with\\\_Explainer}
			&\multicolumn{1}{p{2.0cm}}{\centering Acc\_Global\_\\LocalModel} 
			&\multicolumn{1}{p{2.2cm}}{\centering Acc\_Explainer and \\ Local Model} 
			& \multicolumn{1}{p{2.2cm}}{\centering Acc\_Explainer and \\Global Model}\\ 
			\hline
			NN5W& PROPHET & \textbf{5.601E-18} & \textbf{1.220E-06 }& 2.123E-03 & 1.447E-01 & 4.318E-01 & 8.990E-01 \\ 
			& DHR\_ARIMA & \textbf{5.788E-24} & \textbf{2.087E-09} & 4.433E-04 & 1.883E-01 & 2.640E-01 & 6.748E-01 \\ 
			& TBATS & \textbf{2.693E-23} & \textbf{1.843E-10} & 2.421E-04 & \textbf{1.942E-08} & \textbf{6.629E-06} & 9.943E-01 \\ 
			AusGridW & ETS & 4.203E-04 & \textbf{2.276E-11} & 1.000E+00 & \textbf{1.520E-13} & \textbf{8.872E-05} & 1.000E+00 \\ 
			& PROPHET & \textbf{2.959E-11} & \textbf{5.273E-08} & 1.000E+00 & \textbf{5.919E-05} & \textbf{1.181E-05} & 9.834E-01 \\ 
			& DHR\_ARIMA & \textbf{3.708E-16 }& \textbf{2.295E-14 }& 1.000E+00 & 5.523E-04 & \textbf{5.674E-06} & 8.919E-01 \\ 
			&TBATS & \textbf{3.970E-17} & \textbf{7.418E-05} & 9.973E-01 & \textbf{1.671E-05} & 2.668E-02 & 9.988E-01 \\ 
			& STL\_ETS & \textbf{1.917E-13} & \textbf{3.732E-28} & 9.848E-01 & \textbf{1.564E-09} & \textbf{5.238E-14} & 9.865E-01 \\ 
			WebTrafficD & ETS &\textbf{ 4.714E-22} & \textbf{8.557E-05} & \textbf{1.595E-04} & 1.609E-02 & 1.629E-02 & 6.384E-01 \\ 
			& THETA & \textbf{2.762E-22} & \textbf{7.716E-27 }& \textbf{1.104E-24} &\textbf{ 3.261E-07} & \textbf{4.623E-07} & 9.452E-01 \\ 
			SFTrafficH & TBATS &\textbf{ 9.907E-108} &\textbf{ 2.046E-113} & \textbf{1.157E-92} & \textbf{1.916E-22} & \textbf{1.037E-24 }& 7.013E-02 \\ 
			& PROPHET & \textbf{2.247E-44} & \textbf{5.228E-161} &\textbf{ 1.583E-63} & \textbf{3.434E-195} & \textbf{5.045E-131} & 1.000E+00 \\ 
			& DHR\_ARIMA & \textbf{2.122E-63} & \textbf{1.974E-162} & \textbf{9.674E-57} & \textbf{7.890E-54} &\textbf{ 4.229E-72} & 3.098E-02 \\ 
			& MSTL\_ETS & 5.427E-02 & \textbf{6.423E-71} & \textbf{1.327E-26} & \textbf{1.091E-65} & \textbf{1.171E-21} & 1.000E+00 \\ 
			AusGridHH &TBATS & \textbf{6.712E-09 }& \textbf{9.007E-29 }& \textbf{1.024E-23} & \textbf{3.758E-14} & \textbf{1.045E-16} & 1.871E-01 \\ 
			& PROPHET & \textbf{3.228E-05} & \textbf{3.803E-22} & \textbf{1.488E-21} & \textbf{9.039E-50} & \textbf{4.749E-41} & 1.000E+00 \\ 
			& DHR\_ARIMA & \textbf{8.148E-07 }& \textbf{4.004E-06} & 2.162E-02 & \textbf{9.206E-24 }& \textbf{4.065E-19 }& 9.085E-01 \\ 
			& MSTL\_ETS & 8.287E-01 & \textbf{1.346E-04} & \textbf{1.448E-13} & \textbf{6.323E-33} & 1.209E-02 & 1.000E+00 \\ 
			\hline
		\end{tabular}
	}
\end{table}

\begin{table}[H]
	\centering
	\caption{Unadjusted $p$-values for Mean MAE measures of the datasets for different local explainer models. Significant values that are significant according to the Bonferroni-corrected significance level of $1.6\times 10^{-4}$ are shown in boldface.} 
	\label{tab:mean_mae_stat}
	\scalebox{0.7}{
		\begin{tabular}{llcccccc}
			\hline
			Dataset 
			&\multicolumn{1}{p{1.5cm}}{\centering  Local\\ Explainer} 
			& Fidelity\_Actual 
			& Fidelity\_Local 
			& \multicolumn{1}{p{2.0cm}}{\centering Fidelity\_with\\\_Explainer}
			&\multicolumn{1}{p{2.0cm}}{\centering Acc\_Global\_\\LocalModel} 
			&\multicolumn{1}{p{2.2cm}}{\centering Acc\_Explainer and \\ Local Model} 
			& \multicolumn{1}{p{2.2cm}}{\centering Acc\_Explainer and \\Global Model}\\ 
			\hline
			NN5W& PROPHET & \textbf{5.404E-14 }& \textbf{4.141E-07} & 3.266E-03 & 1.445E-01 & 4.007E-01 & 9.192E-01 \\ 
			& DHR\_ARIMA & \textbf{1.002E-19} & \textbf{2.037E-07} & 6.889E-04 & 1.608E-01 & 2.058E-01 & 6.853E-01 \\ 
			&  TBATS & \textbf{1.510E-19} & \textbf{1.011E-10} & 2.219E-04 & \textbf{1.570E-07} & \textbf{8.564E-06} & 9.944E-01 \\ 
			AusGridW & ETS & 1.848E-04 & \textbf{5.806E-10} & 1.000E+00 & \textbf{6.092E-12} & 5.061E-04 & 1.000E+00 \\ 
			& PROPHET & \textbf{2.293E-07} & \textbf{1.539E-07} & 1.000E+00 &\textbf{ 4.115E-05} & \textbf{2.117E-05} & 9.956E-01 \\ 
			& DHR\_ARIMA & \textbf{8.960E-13} & \textbf{5.240E-09} & 1.000E+00 & 5.632E-04 & 1.881E-04 & 9.553E-01 \\ 
			& TBATS &\textbf{ 1.877E-14} & 1.230E-03 & 9.911E-01 & \textbf{1.445E-04} & 2.180E-01 & 9.991E-01 \\ 
			& STL\_ETS & \textbf{4.926E-10} & \textbf{3.390E-21} & 9.974E-01 &\textbf{ 5.320E-09} & \textbf{2.228E-12 }& 9.954E-01 \\ 
			WebTrafficD & ETS & \textbf{1.784E-15} & \textbf{2.249E-10} & \textbf{1.665E-10} &\textbf{ 2.581E-05} &\textbf{ 3.233E-05}\textbf{} & 2.908E-01 \\ 
			& THETA & \textbf{2.122E-15} & \textbf{8.827E-06} & \textbf{6.573E-06} & 5.022E-03 & 6.546E-03 & 5.068E-01 \\ 
			SFTrafficH & TBATS & \textbf{3.017E-91} & \textbf{5.435E-117} & \textbf{5.991E-97} & \textbf{1.257E-14} & \textbf{1.635E-12} & 8.229E-01 \\ 
			& PROPHET & \textbf{2.441E-36} & \textbf{3.805E-137} & \textbf{8.329E-57} & \textbf{5.589E-200} & \textbf{7.903E-107} & 1.000E+00 \\ 
			& DHR\_ARIMA & \textbf{1.497E-54} &\textbf{ 4.944E-137} &\textbf{ 4.781E-49} & \textbf{7.527E-52} & \textbf{1.049E-59} & 7.206E-01 \\ 
			& MSTL\_ETS & 5.143E-01 & \textbf{8.950E-59} & \textbf{4.502E-17} & \textbf{2.143E-56} & \textbf{4.498E-11} & 1.000E+00 \\ 
			AusGridHH &TBATS  & \textbf{3.109E-61} & \textbf{1.864E-15} & \textbf{4.808E-13} & \textbf{5.149E-10} &\textbf{ 2.780E-10} & 5.214E-01 \\ 
			& PROPHET & \textbf{4.907E-28} & \textbf{2.156E-38} & \textbf{3.959E-32} & \textbf{4.740E-35} & \textbf{2.390E-30} & 1.000E+00 \\ 
			& DHR\_ARIMA & \textbf{6.519E-48} & \textbf{7.832E-58} & \textbf{5.449E-07} & \textbf{1.955E-22} &\textbf{ 2.485E-27} & 9.996E-01 \\ 
			& MSTL\_ETS & 9.961E-01 & 5.608E-04 & \textbf{5.263E-14 }&\textbf{ 1.292E-21} & 5.488E-03 & 1.000E+00 \\ 
			\hline
		\end{tabular}
	}
\end{table}

\section{Statistical Significance of the Results of NSTL explainer}
\begin{table}[H]
	\centering
	\caption{Unadjusted $p$-values for Mean RMSE measures of the datasets for different local explainer models. Significant values that are significant according to the Bonferroni-corrected significance level of $1.6\times 10^{-4}$ are shown in boldface.} 
	\label{tab:mean_rmse_stat_bm1}
	\scalebox{0.7}{
		\begin{tabular}{llcccccc}
			\hline
			Dataset 
			&\multicolumn{1}{p{1.5cm}}{\centering  Local\\ Explainer} 
			& Fidelity\_Actual 
			& Fidelity\_Local 
			& \multicolumn{1}{p{2.0cm}}{\centering Fidelity\_with\\\_Explainer}
			&\multicolumn{1}{p{2.0cm}}{\centering Acc\_Global\_\\LocalModel} 
			&\multicolumn{1}{p{2.2cm}}{\centering Acc\_Explainer and \\ Local Model} 
			& \multicolumn{1}{p{2.2cm}}{\centering Acc\_Explainer and \\Global Model}\\ 
			\hline
			NN5W& PR & \textbf{9.128E-12} & 7.651E-01 & 3.846E-01 & 5.501E-04 & 3.022E-03 & 8.623E-01 \\ 
			WebTraffic100D & ETS & \textbf{9.445E-16} & \textbf{6.354E-05} &\textbf{ 6.883E-05} & 1.187E-02 & 1.101E-02 & 2.357E-01 \\ 
			& PR & \textbf{6.398E-15} & \textbf{1.518E-05} & \textbf{1.578E-05 }& 5.655E-02 & 4.098E-02 & 4.593E-01 \\ 
			\hline
		\end{tabular}
	}
\end{table}
\begin{table}[H]
	\centering
	\caption{Unadjusted $p$-values for Mean MASE measures of the datasets for different local explainer models} 
	\label{tab:mean_mase_stat_bm1}
	\scalebox{0.7}{
		\begin{tabular}{llcccccc}
			\hline
			Dataset 
			&\multicolumn{1}{p{1.5cm}}{\centering  Local\\ Explainer} 
			& Fidelity\_Actual 
			& Fidelity\_Local 
			& \multicolumn{1}{p{2.0cm}}{\centering Fidelity\_with\\\_Explainer}
			&\multicolumn{1}{p{2.0cm}}{\centering Acc\_Global\_\\LocalModel} 
			&\multicolumn{1}{p{2.2cm}}{\centering Acc\_Explainer and \\ Local Model} 
			& \multicolumn{1}{p{2.2cm}}{\centering Acc\_Explainer and \\Global Model}\\ 
			\hline
			NN5W & PR & \textbf{6.164E-06} & 9.922E-01 & 9.531E-01 & \textbf{1.409E-04} & 2.856E-02 & 9.374E-01 \\ 
			WebTraffic100D & ETS & \textbf{8.372E-12} & \textbf{2.166E-05} & \textbf{1.901E-05} & 1.059E-03 & 3.423E-03 & 9.987E-01 \\ 
			& PR & \textbf{2.250E-11} & 9.054E-04 & \textbf{3.868E-06} & 7.678E-04 & 1.913E-03 & 2.175E-01 \\ 
			\hline
		\end{tabular}
	}
\end{table}

\begin{table}[H]
	\centering
	\caption{Unadjusted $p$-values for Mean MAE measures of the datasets for different local explainer models. Significant values that are significant according to the Bonferroni-corrected significance level of $1.6\times 10^{-4}$ are shown in boldface.} 
	\label{tab:mean_mae_stat_bm1}
	\scalebox{0.7}{
		\begin{tabular}{llcccccc}
			\hline
			Dataset 
			&\multicolumn{1}{p{1.5cm}}{\centering  Local\\ Explainer} 
			& Fidelity\_Actual 
			& Fidelity\_Local 
			& \multicolumn{1}{p{2.0cm}}{\centering Fidelity\_with\\\_Explainer}
			&\multicolumn{1}{p{2.0cm}}{\centering Acc\_Global\_\\LocalModel} 
			&\multicolumn{1}{p{2.2cm}}{\centering Acc\_Explainer and \\ Local Model} 
			& \multicolumn{1}{p{2.2cm}}{\centering Acc\_Explainer and \\Global Model}\\ 
			\hline
			NN5W & PR  & \textbf{2.454E-05 }& 7.897E-01 & 3.816E-01 & 2.299E-04 & 1.254E-03 & 8.786E-01 \\ 
			WebTraffic100D & ETS & \textbf{4.056E-16} & \textbf{7.766E-05} & \textbf{7.929E-05} & 2.698E-03 & 2.625E-03 & 2.652E-01 \\ 
			& PR & \textbf{3.172E-15 }& \textbf{6.637E-06} & \textbf{1.132E-05} & \textbf{7.251E-05} & 3.468E-04 & 7.666E-01 \\ 
			\hline
		\end{tabular}
	}
\end{table}

\section{Statistical Significance of the Results of NSieve explainer}

\begin{table}[H]
	\centering
	\caption{Unadjusted $p$-values for Mean RMSE measures of the datasets for different local explainer models. Significant values that are significant according to the Bonferroni-corrected significance level of $1.6\times 10^{-4}$ are shown in boldface.} 
	\label{tab:mean_rmse_stat_bm2}
	\scalebox{0.7}{
		\begin{tabular}{llcccccc}
			\hline
			Dataset 
			&\multicolumn{1}{p{1.5cm}}{\centering  Local\\ Explainer} 
			& Fidelity\_Actual 
			& Fidelity\_Local 
			& \multicolumn{1}{p{2.0cm}}{\centering Fidelity\_with\\\_Explainer}
			&\multicolumn{1}{p{2.0cm}}{\centering Acc\_Global\_\\LocalModel} 
			&\multicolumn{1}{p{2.2cm}}{\centering Acc\_Explainer and \\ Local Model} 
			& \multicolumn{1}{p{2.2cm}}{\centering Acc\_Explainer and \\Global Model}\\ 
			\hline
			NN5W & PR  & \textbf{7.741E-10} & 8.430E-01 & 9.999E-01 & 5.501E-04 & 4.664E-01 & 1.000E+00 \\ 
			WebTraffic100D & ETS & \textbf{4.312E-15} & 6.938E-04 & \textbf{6.604E-05} & 1.187E-02 & 3.467E-02 & 9.999E-01 \\ 
			& PR &\textbf{ 3.185E-13 }& 3.882E-01 & 7.884E-04 & 5.399E-02 & 8.547E-01 & 9.998E-01 \\ 
			\hline
		\end{tabular}
	}
\end{table}

\begin{table}[H]
	\centering
	\caption{Unadjusted $p$-values for Mean MASE measures of the datasets for different local explainer models. Significant values that are significant according to the Bonferroni-corrected significance level of $1.6\times 10^{-4}$ are shown in boldface.} 
	\label{tab:mean_mase_stat_bm2}
	\scalebox{0.7}{
		\begin{tabular}{llcccccc}
			\hline
			Dataset 
			&\multicolumn{1}{p{1.5cm}}{\centering  Local\\ Explainer} 
			& Fidelity\_Actual 
			& Fidelity\_Local 
			& \multicolumn{1}{p{2.0cm}}{\centering Fidelity\_with\\\_Explainer}
			&\multicolumn{1}{p{2.0cm}}{\centering Acc\_Global\_\\LocalModel} 
			&\multicolumn{1}{p{2.2cm}}{\centering Acc\_Explainer and \\ Local Model} 
			& \multicolumn{1}{p{2.2cm}}{\centering Acc\_Explainer and \\Global Model}\\ 
			\hline
			NN5W & PR & \textbf{1.210E-08} & 9.985E-01 & 1.000E+00 & \textbf{1.409E-04} & 9.452E-01 & 1.000E+00 \\ 
			WebTraffic100D & ETS & \textbf{6.209E-12} & 1.141E-03 & \textbf{2.401E-05} & 3.426E-03 & 2.092E-02 & 9.983E-01 \\ 
			& PR & \textbf{1.685E-11} & 3.919E-01 & 6.182E-03 & 6.457E-04 & 4.506E-01 & 9.966E-01 \\ 
			\hline
		\end{tabular}
	}
\end{table}
\begin{table}[H]
	\centering
	\caption{Unadjusted $p$-values for Mean MAE measures of the datasets for different local explainer models. Significant values that are significant according to the Bonferroni-corrected significance level of $1.6\times 10^{-4}$ are shown in boldface.} 
	\label{tab:mean_mae_stat_bm2}
	\scalebox{0.7}{
		\begin{tabular}{llcccccc}
			\hline
			Dataset 
			&\multicolumn{1}{p{1.5cm}}{\centering  Local\\ Explainer} 
			& Fidelity\_Actual 
			& Fidelity\_Local 
			& \multicolumn{1}{p{2.0cm}}{\centering Fidelity\_with\\\_Explainer}
			&\multicolumn{1}{p{2.0cm}}{\centering Acc\_Global\_\\LocalModel} 
			&\multicolumn{1}{p{2.2cm}}{\centering Acc\_Explainer and \\ Local Model} 
			& \multicolumn{1}{p{2.2cm}}{\centering Acc\_Explainer and \\Global Model}\\ 
			\hline
			NN5W & PR & \textbf{2.001E-05 }& 8.874E-01 & 9.999E-01 & 2.299E-04 & 4.477E-01 & 1.000E+00 \\ 
			WebTraffic100D & ETS & \textbf{9.968E-16} & 1.262E-03 & \textbf{7.991E-05} & 2.698E-03 & 8.099E-03 & 9.861E-01 \\ 
			& PR & \textbf{4.924E-11} & 5.086E-01 & 2.110E-03 & \textbf{7.423E-05 }& 1.389E-01 & 9.966E-01 \\ 
			\hline
		\end{tabular}
	}
\end{table}

\end{document}